\documentclass[10pt]{article} 
\usepackage[accepted]{tmlr}

\usepackage{amsmath,amsfonts,bm}

\def\eqref#1{equation~\ref{#1}}

\def\1{\bm{1}}

\DeclareMathAlphabet{\mathsfit}{\encodingdefault}{\sfdefault}{m}{sl}
\SetMathAlphabet{\mathsfit}{bold}{\encodingdefault}{\sfdefault}{bx}{n}

\usepackage{hyperref}
\usepackage{url}
\usepackage[utf8]{inputenc} 
\usepackage[T1]{fontenc}    
\usepackage{booktabs}       
\usepackage{amsfonts}       
\usepackage{nicefrac}       
\usepackage{microtype}      
\usepackage{xcolor}         
\usepackage{graphicx}
\usepackage{subfig}
\usepackage{natbib}
\usepackage{amsmath}
\usepackage{algorithm}
\usepackage{algpseudocode}
\usepackage{cancel}

\title{Emergent Symbol-like Number Variables in Artificial Neural Networks}

\author{\name Satchel Grant \email grantsrb@stanford.edu \\
      \addr Departments of Psychology and Computer Science \\
      Stanford University
      \AND
      \name Noah D. Goodman \email ngoodman@stanford.edu \\
      \addr Departments of Psychology and Computer Science \\
      Stanford University
      \AND 
      \name James L. McClelland \email jlmcc@stanford.edu \\
      \addr Departments of Psychology and Computer Science \\
      Stanford University}

\def\month{07}  
\def\year{2025} 
\def\openreview{\url{https://openreview.net/forum?id=YPnYpiru5W}} 

\begin{document}

\newcommand{\dmodel}{128}
\newcommand{\nepochs}{800}
\newcommand{\batchsize}{128}
\newcommand{\maxcount}{20}
\newcommand{\holdouts}{4, 9, 14, and 17}
\newcommand{\hidsize}{512}
\newcommand{\dropout}{0.5}
\newcommand{\nonlinearity}{GELU}
\newcommand{\learnrate}{0.0001}
\newcommand{\ntrain}{1,000}
\newcommand{\nval}{500}
\newcommand{\nseeds}{5}
\newcommand{\seqlen}{43} 

\newcommand{\ndemotypes}{3}
\newcommand{\singleobject}{Single-Object}
\newcommand{\sameobject}{Same-Object}
\newcommand{\multiobject}{Multi-Object}
\newcommand{\varylen}{Variable-Length}
\newcommand{\varyprob}{0.2}
\newcommand{\arithmaxnew}{9}
\newcommand{\arithnumbase}{10}
\newcommand{\aritmaxcumu}{100}

\newcommand{\countupdown}{Up-Down}
\newcommand{\countupup}{Up-Up}
\newcommand{\incrupup}{Increment-Up}
\newcommand{\varcount}{Count}
\newcommand{\varphase}{Phase}
\newcommand{\vardemocount}{Demo Count}
\newcommand{\varrespcount}{Resp Count}
\newcommand{\varlastval}{Input Value}
\newcommand{\varprogress}{Progress}
\newcommand{\varincrement}{Increment}
\newcommand{\varinterval}{Interval}
\newcommand{\fulldistrsoln}{Context Distributed}
\newcommand{\distrsoln}{Ctx-Distr}
\newcommand{\inptval}{Tok Val}
\newcommand{\tokhist}{Tok History}
\newcommand{\eosreadout}{EOS Readout}

\newcommand{\dasntrain}{10,000}
\newcommand{\dasnval}{1,000}
\newcommand{\dasbatchsize}{512}
\newcommand{\daslr}{0.001}
\newcommand{\dasnepochs}{1,000}
\newcommand{\dasdims}{16}

\newcommand{\MultiObjTransformerCtxDistrDASMean}{0.964}
\newcommand{\VaryLenTransformerCtxDistrDASMean}{0.949}

\maketitle

\begin{abstract}
What types of numeric representations emerge in neural systems, and
what would a satisfying answer to this question look like?
In this work, we interpret Neural Network (NN) solutions to sequence
based number tasks through a variety of methods to
understand how well we can interpret them through the lens of interpretable
Symbolic Algorithms (SAs)---precise algorithms describable by rules operating
on typed, mutable variables. We use GRUs, LSTMs, and Transformers trained using
Next Token Prediction (NTP) on tasks where the correct tokens depend
on numeric information only latent in the task structure. We show through
multiple causal and theoretical methods that we can interpret raw NN activity
through the lens of simplified SAs when we frame the neural activity in terms
of neural subspaces rather than individual neurons. Using Distributed Alignment
Search (DAS), we find that, depending on network architecture, dimensionality,
and task specifications, alignments with SA's can be very high,
while other times they can be only approximate, or fail altogether. We extend
our analytic toolkit to address the failure cases by expanding the DAS framework
to a broader class of alignment functions that more flexibly capture NN activity
in terms of interpretable variables from SAs, and we provide theoretic and
empirical explorations of Linear Alignment Functions (LAFs) in contrast to the
preexisting Orthogonal Alignment Functions (OAFs). Through
analyses of specific cases we confirm the usefulness of causal interventions
on neural subspaces for NN interpretability, and we show that recurrent models
can develop graded, symbol-like number variables within their neural activity.
We further show that shallow Transformers learn very different solutions than
recurrent networks, and we prove that such models must use anti-Markovian
solutions---solutions that do not rely on cumulative, Markovian hidden
states---in the absence of sufficient attention layers.
\end{abstract}

\section{Introduction}

We can see examples of the modeling power of Neural Networks (NNs) in
both biological NNs (BNNs) from the impressive capabilities of
human cognition, and in artificial NNs (ANNs) where recent advances
have had such great success that ANNs have been crowned the “gold standard”
in many machine learning communities \citep{Alzubaidi2021DLReview}.
The inner workings of NNs, however, are still often opaque. This is, in
part, due to their representations being highly distributed. Individual
neurons can play multiple roles within a network in what's called population
encoding. In these cases, human-interpretable information is encoded across
populations of neurons rather than within any individual unit
\citep{hinton1986distributed, Smolensky1988, olah2017feature,olah2020zoom,elhage2022superposition, scherlis2023polysemanticity,olah2023compositionsuperposition}.

Symbolic Algorithms (SAs), in contrast, defined as processes that
manipulate distinct, typed entities according to explicit rules
and relations, can have the benefit of consistency, transparency, and
generalization when compared to their neural counterparts. A concrete example of
an SA is a computer program,
where the variables are distinct, typed, mutable entities, able to represent many
different values, processed by well defined functions.
There are many existing theories that posit the necessity of algorithmic,
symbolic, processing for higher level cognition
\citep{do2021neuralsymbols,fodor1988connectionism,fodor1975,fodor1987,newell1980,newell1982,pylyshyn1980,marcus2018deeplearningcriticalappraisal,lake_building_2017}.
Human designed symbolic cognitive systems, however, can lack the expressivity
and performance of NNs. This is
apparent in the field of natural language processing where neural architectures
trained on vast amounts of data \citep{vaswani2017,brown2020gpt3,kaplan2020scaling}
have swept the field, surpassing the pre-existing symbolic approaches.
Despite the differences between NNs and SAs, it might be argued that
NNs actually implement simplified SAs; or, they may approximate
them well enough that seeking neural analogies to these simplified SAs
would be a powerful step toward an interpretable, unified understanding
of complex neural behavior. In one sense, this pursuit is trivial for ANNs,
in that ANNs are, by definition, aligned to the computer program that
defines them.
The algorithms they acquire through training, however, 
are not so trivial to understand, and their complexity can be so great
that simplified SAs become useful for explaining, predicting, and controlling
their behavior. This approach of seeking to characterize NNs (either
biological or artificial) in terms of simplified, interpretable SAs is one
way of construing the goal of many research programs in cognitive
science, neuroscience, and mechanistic interpretability.
\begin{figure}[t!]
  \centering
  \includegraphics[width=0.9\textwidth]{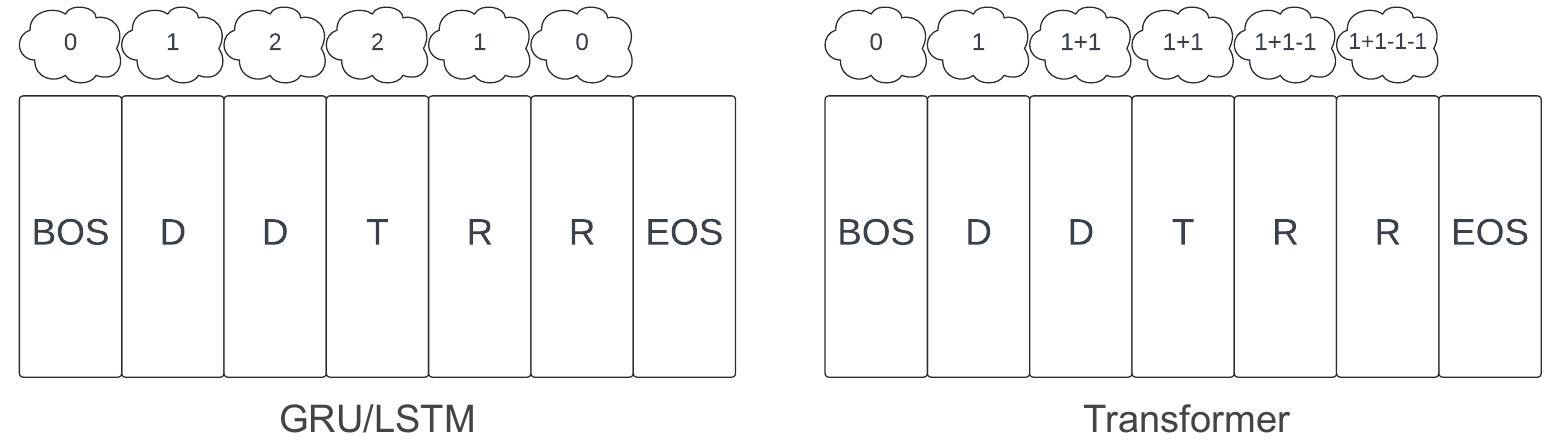}
  \caption{
        Different architecture's solutions achieving the same
        accuracy on a numeric equivalence task. The rectangles represent
        tokens for a task in which the model must produce the same number
        of R tokens ending with the EOS token as it observed D tokens. The
        T token indicates the end of the D tokens (see Methods~\ref{sec:tasks}).
        The thought bubbles represent the values of causally discovered neural variables
        encoded within the models' representations. The recurrent models encode a
        single count variable that increments up before the T token and down after the T
        token, with 0 indicating the end of the task.
        Transformers learn a solution in which they recompute the task relevant
        information from the input tokens at each step in the sequence.
        All NoPE transformers align with the displayed Transformer solution.
        RoPE transformers can partially rely on positional information unless they
        are trained on a variant of the task that breaks number-position correlations.
        }
    \label{fig:solutions}
\end{figure}

In this work, we narrow our focus to numeric cognition and ask, how can we
understand neural implementations of numeric values at the level of SAs?
Numeric reasoning has the advantage of being well studied in humans of
different ages and experience levels, which provides a powerful domain for
comparisons between BNNs and ANNs \citep{Nuovo2019numcogreview}. And
numeric domains provide the benefit of tasks built upon well defined variables.
We focus on a numeric equivalence task that was used to test the numeric
abilities of humans whose language lacks explicit number words \citep{gordon2004}.
The task is formulated as a sequence of tokens, requiring the subject to produce
the same number of response tokens as a quantity of demonstration tokens initially
observed at the beginning of the task. This
task is interesting for computational settings because the training labels vary
in both identity and sequence length, and quantities are never explicitly labeled.
Similar versions of this task have also been used in previous theoretical
and computational work \citep{el-naggar2023rnncounting,weiss2018GRUsLSTMscounting,behrens2024transformercount},
providing a platform to expand our understanding of seemingly disparate
modeling systems in unified ways.

What sorts of representations do ANNs use to solve such a task and how do they
arrive at these representations? Do the networks represent numbers in a shared
system, or do they use different systems for different situations? Is it
propitious to think about their representations as though they are discrete
variables in an SA, or would it be better to think of their neural activity on
a graded continuum?
Do the answers to these questions change over the course of training, and do the
answers vary based on task and architectural details? How can we unify the way
we understand NN solutions in satisfying ways for cognitive scientists,
neuroscientists, and computer scientists alike? We set out to understand
NN neural activity through the lens of simplified, interpretable SAs using
causal interventions to support our interpretations.

In this work, we pursue these questions by training Gated Recurrent Units
(GRUs)\citep{Cho2014GRU}, Long Short-Term Memory cells (LSTMs)
\citep{hochreiter1997LSTM}, and Transformers on numeric equivalence tasks using
Next Token Prediction (NTP).
We then provide causal, correlational, and theoretical analyses such as
activation patching, Principal Component Analysis (PCA), attention visualizations,
and Distributed Alignment Search (DAS) \citep{geiger2021,geiger2023boundless} to
understand the networks' representations and solutions, and we introduce the
notion of an Alignment Function to the DAS framework which we use to frame neural
activity in terms of functions of interpretable variables. We summarize
our contributions as follows:

\hangindent=0.5cm 1. We show through causal interventions the emergence
of graded symbol-like neural variables in RNNs, where neural variables are defined
as representational subspaces that causally align with a variable from
an SA, and they are "graded" or "symbol-like" when they exhibit signatures of
a continuum rather than being fully discrete.

\hangindent=0.5cm 2. We show that seemingly insignificant task variations
can drastically affect the NN's alignment to the SAs, motivating us to
extend the DAS framework to a broader class of alignment functions,
including linear alignment functions (LAFs). We show that LAFs can be used
to control NN behavior through their neural activity aligned to
interpretable SAs, and we provide a theoretical
analysis of LAFs and Orthogonal Alignment Functions (OAFs) to better
understand their properties.


\hangindent=0.5cm 3. We show empirically that shallow Transformers use an
anti-Markovian solution to the numeric tasks, and we show theoretically
that Transformers must use anti-Markovian solutions in all tasks in the
absence of sufficient attention layers.

\hangindent=0.5cm 4. Through our specific analyses, we confirm the
utility of interpreting NNs at the level of neural subspaces
and using causal interventions to make claims about NN solutions.

We use our results to encourage the use of multiple causal interpretability
tools for any representational analysis, to highlight functional differences
that might emerge from architectural/task constraints, and to demonstrate the
nuances of interpreting neural activity at the level of SAs.

\section{Related Work}
Many prior works have attempted to describe ANNs through SAs. Some
relevant examples include
\citet{lindner2023tracrcompiledtransformers} who built a system to compile
human written code into transformer models and
\citet{michaud2024openingaiblackbox} who perform linear and symbolic
regression on simplified NN representations to generate programs that
perform like the NNs. Both of these cases focus on generating sufficient
SAs for network behavior rather than relating the internal mechanisms
of an existing NN to the resulting SAs. Another prominent example is the work of
\citet{nanda2023progressmeasuresgrokking} who showed that small transformers
trained on modular addition use discrete Fourier transforms
combined with trig identities to solve the task. Nanda et al. used a
combination of examinations and ablations to defend their claims.

We wish to highlight the importance of using causal manipulations for
interpreting NN solutions rather than relying on correlational
analyses. Causal inference broadly refers to processes that isolate causal
effects of individual components within a larger system
\citep{pearl2010causalmediation}. An abundance of causal interpretability
variants have been used to determine what functions are being performed by NN
models' activations and circuits
\citep{olah2018buildingblocks,olah2020zoom,wang2022patching,geva2023dissectingrecallfactual,merrill2023talecircuitsgrokking,bhaskar2024heuristiccore,wu2024alpacadas}.
\citet{vig2020causalmediationanalysis} provides an integrative review of
the rationale for and utility of causal mediation in NN analyses.
We rely and build upon DAS for many of our analyses. DAS can be
thought of as a specific
type of activation patching (also referred to as causal tracing)
\citep{meng2023activpatching,vig2020causalmediationanalysis,heimersheim2024interpretactivationpatching}.
In section 3.4, we introduce the notion of generalized
\emph{alignment functions} which can be thought of as a specific class
of mapping models from \citet{ivanova2022mappingmodels} where the
mapping model in our case defines an invertible, causal relationship
between neural subspaces from an NN and interpretable variables from an
SA; the success of the relationship is measured through NN behavior. Other
works such as \citet{williams2021shapemetrics} have explored differences
between orthogonal and linear relationships between distributed
representations where they largely focused their analyses on formally
defining metrics on representational dissimilarity.

Many publications explore ANNs' abilities to perform counting tasks
\citep{DiNuovo2019,fang2018,sabathiel2020,Kondapaneni2020,Nasr2019,Zhang2018a,Trott2018a}
and closely related tasks \citep{csordás2024}.
Our tasks and modeling paradigms differ from many of these publications
in that numbers are only latent in the structure of our tasks
without explicit teaching of distinct symbols for distinct numeric values.
\citet{el-naggar2023rnncounting} provided a theoretical treatment of
Recurrent Neural Network (RNN) solutions to a parentheses closing task, and
\citet{weiss2018GRUsLSTMscounting} explored Long Short-Term Memory RNNs
(LSTMs) \citep{hochreiter1997LSTM} and Gated Recurrent Units (GRUs)
\citep{Cho2014GRU} in a similar numeric equivalence task looking at the
activations. These works showed correlates of a magnitude scaling solution
in both theoretical and practically trained ANNs. Our work builds on their
findings by using causal methods for our analyses, expanding the models
considered, and introducing new theoretical and empirical analyses.
\citet{behrens2024transformercount}
explored transformer counting solutions in a task similar to ours. Our work
extends beyond theirs by exploring positional encodings, avoiding explicit
labels of the numeric concepts, using causal analyses, and providing
a theoretical focus on anti-Markovian solutions.
\section{Methods}
\subsection{Numeric Equivalence Tasks}\label{sec:tasks}
Each task we consider is defined by a Next-Token Prediction (NTP) task
over sequences, as in the example shown in Figure~\ref{fig:solutions}.
The goal of the task is to reproduce the same number of response
tokens as demonstration tokens observed before the Trigger (T) token.
Each sequence starts with a Beginning of Sequence (BOS) token
and ends with an End of Sequence (EOS) token. Each sequence is
generated by first uniformly sampling an object quantity
from the inclusive range of 1 to \maxcount\space where \maxcount\space
was chosen to match the human experiments of \citet{pitt2022}.
The sequence is then
constructed as the combination of two phases. The first phase, called
the demonstration phase (\textbf{demo phase}),
starts with the BOS token and continues with a series of demo
tokens equal in quantity to the sampled object quantity. The end of
the demo phase is indicated by the trigger token after the
demo tokens. This also marks the beginning of the response phase (\textbf{resp phase}).
The resp phase consists of a series of resp
tokens equal in number to the demo tokens. After the resp tokens,
the end of the sequence is denoted by the EOS token.

During the teacher forced, NTP model training, we include all tokens in the
NTP loss. During model evaluation and DAS trainings,
we only consider tokens in the resp phase---which are fully determined by
the demo phase. During model trainings, we hold out the object quantities
\holdouts\space as a way to examine generalization. We chose \holdouts\space
to semi-uniformly cover the space of possible numbers while including even,
odd, and prime numbers and ensuring that training covered 3 examples at
both ends of the training range. A trial is considered correct when all
resp tokens and the EOS token are correctly predicted by the model after the
trigger. We include three variants of this task differing only in their demo
and resp token instances.

\textbf{\multiobject\space Task:} there are 3
demo token instances \{$\text{D}_1$, $\text{D}_2$, $\text{D}_3$\}
with a single response token instance, R. The demo tokens are uniformly
sampled from the 3 possible instances. An example input sequence
with an object quantity of 2 could be: "BOS $\text{D}_3$ $\text{D}_1$ T",
with a ground truth response of "R R EOS".  All possible tokens are
contained in the set
$\{\text{BOS, D}_1\text{, D}_2\text{, D}_3\text{, T, R, EOS}\}$.
    
\textbf{\singleobject\space Task:} there is a single
demo token instance, D, and a single response token instance, R.
An example of the input sequence with an object quantity of 2 is:
"BOS D D T", with a ground truth response of "R R EOS". All
possible tokens are contained in the set
$\{\text{BOS, D, T, R, EOS}\}$.
    
\textbf{\sameobject\space Task:} there is a single
token instance, C, used for both the demo and resp tokens. An
example of the input sequence with an object quantity of 2 is:
"BOS C C T", with a ground truth response of "C C EOS".  All
possible tokens are contained in the set $\{\text{BOS, C, T, EOS}\}$.

For some transformer trainings, we include \textbf{\varylen\space (VL)} variants of each
task to break count-position correlations.
In these variants, each token in the demo phase has a \varyprob\space probability
of being sampled as a unique
"void" token type, V, that should be ignored when determining
the object quantity of the sequence. The number of demo tokens
will still be equal to the object quantity when the trigger token
is presented.
As an example, consider the possible sequence with an object quantity
of 2: "BOS V D V V D T R R EOS".

\subsection{Model Architectures}
The recurrent models in this paper consist of Gated Recurrent
Units (GRUs) \citep{Cho2014GRU}, and Long Short-Term Memory
networks (LSTMs) \citep{hochreiter1997LSTM}. These architectures
both have a Markovian, hidden state vector that bottlenecks
all predictive computations following the structure:
\begin{equation}
h_{t+1} = f(h_t, x_t) \\
\end{equation}
\begin{equation}
\hat{x}_{t+1} = g(h_{t+1})\\
\end{equation}
Where $h_t$ is the hidden state vector at step $t$, $x_t$ is the input
token at step $t$, $f$ is the recurrent function (either a GRU or LSTM
cell), and $g$ is a multi-layer perceptron (MLP) used to make a
prediction, denoted $\hat{x}_{t+1}$, of the token at step $t+1$. 

We contrast the recurrent architectures against transformer architectures
\citep{vaswani2017,touvron2023llama,su2023roformer} in that the
transformers use a history of input tokens,
$X_t = [x_1, x_2, ..., x_t]$, at each time step, $t$, to make a
prediction:
\begin{equation}
\hat{x}_{t+1} = f(X_t) \\
\end{equation}
Where $f$ now represents the transformer architecture. We show results
from 2 layer, single attention head transformers that use No
Positional Encodings (NoPE) \citep{haviv2022NoPE} and Rotary Positional Encodings (RoPE)
\citep{su2023roformer}.
Refer to Supplemental Figure~\ref{fig:tformerarch}
for more model and architectural details. We also consider one-layer
transformers
with No Positional Encodings (NoPE) in Results section~\ref{sec:resultsNoPE}.
For all of our analyses except the training curves in
Figure~\ref{fig:sizegradience},
we first train the models to $>99\%$ accuracy on their respective
tasks before performing analyses.
The models are evaluated on 15 sampled sequences of each of
the 16 trained and 4 held out object quantities. We train \nseeds\space
model seeds for each training condition. One seed from
the transformer models in both the \varylen\space \multiobject\space and
\varylen\space \sameobject\space tasks were dropped without replacement
due to their low accuracy.

\subsection{Symbolic Algorithms (SAs)}
In this work, we examine the alignment of 3 different SAs
to the models' distributed representations.

\textbf{\countupdown\space Program:} uses a single numeric
variable, called the \textbf{\varcount}, to track the difference
between the number of demo tokens and resp tokens at each step
in the sequence. It also contains a \textbf{\varphase} variable
to determine whether it is in the demo or resp phase.
The program ends when the \varcount\space is equal to 0 during the
resp phase.
    
\textbf{\countupup\space Program:} uses two numeric
variables---the \textbf{\vardemocount\space} and
\textbf{\varrespcount}---to track quantities at each step in
the sequence. It uses a \varphase\space variable to track
which phase it is in.
This program increments the \vardemocount\space during
the demo phase and increments the \varrespcount\space during the
resp phase. It ends when the \vardemocount\space is equal to the
\varrespcount\space during the resp phase.
    
\textbf{\fulldistrsoln\space (\distrsoln) Program:}
queries a history of inputs at each step in the
sequence, assigns a numeric value to each, and sums their values
to determine when to stop (contrasted against encoding a cumulative,
Markovian quantity variable). More specifically, this program uses
an \varlastval\space variable for each input token, and assigns the
\varlastval\space a value of 1 for demo tokens and -1 for resp tokens
and computes the sum of the \varlastval s at each step
in the sequence to determine the count. This program outputs the EOS
token when the sum is 0 and the sequence contains the T token.

We include Algorithms~\ref{alg:updown},~\ref{alg:upup}, and ~\ref{alg:distr}
in the supplement which show the pseudocode used to implement the
\countupdown, \countupup, and \distrsoln\space programs in simulations.
Refer to Figure~\ref{fig:solutions} for an illustration of the
\countupdown\space strategy and the \distrsoln\space strategy that is
observed in some transformers.

It is important to note that there are an infinite number of causally equivalent
implementations of these SAs. For example, the \countupdown\space
program could immediately add and subtract 1 from the \varcount\space
at every step of the task in addition to carrying out the rest of the program as
previously described. We do not discriminate between programs that are causally
indistinct from one another in this work.

\subsection{Distributed Alignment Search (DAS)}\label{meth:das}
%
DAS measures the degree of alignment between a representational
subspace from an NN and a symbolic variable from a symbolic algorithm (SA) by
testing the assumption that the model hidden state $h \in R^{d_m}$
can be written as an orthogonal rotation $z = Qh$, where $Q\in R^{d_m\times d_m}$
is orthonormal, $z \in R^{d_m}$ consists of contiguous subspaces encoding
high-level variables from SAs, and $d_m$ is the size of the hidden state.
The benefit of this alignment is that it allows us to understand the NN's activity
through interpretable variables and it allows us to manipulate the value of
these variables without affecting other information.

Concretely, DAS performed on the \countupdown\space program tests the
hypothesis that $z$ is composed of subspaces $\Vec{z}_{\text{count}}$
encoding the \varcount, $\Vec{z}_{\text{phase}}$ encoding the \varphase, and
$\Vec{z}_{\text{extra}}$ encoding extraneous, irrelevant activity.
\begin{eqnarray}\label{eq:zdef}
    z = \begin{bmatrix}
        \Vec{z}_{\text{count}} \\
        \Vec{z}_{\text{phase}} \\
        \Vec{z}_{\text{extra}}
    \end{bmatrix}
\end{eqnarray}
Each $\Vec{z}_{\text{var}} \in R^{d_{\text{var}}}$ is a column vector of
potentially different lengths satisfying the relation
$d_{\text{count}} + d_{\text{phase}} + d_{\text{extra}} = d_m$.
Under this assumption, the value of a high-level variable encoded in $h$
can be freely exchanged through causal interventions using:
\begin{equation}\label{eqn:interchange}
    h^v = Q^{-1}((1-D_{\text{var}})Qh^{trg} + D_{\text{var}}Qh^{src})
\end{equation}
Where $D_{\text{var}}\in R^{d_{m}\times d_{m}}$ is a manually chosen,
diagonal, binary matrix with $d_{\text{var}}$ non-zero elements
used to isolate the dimensions that make up $\Vec{z}_{\text{v}ar}$,
$h^{src}$ is the \emph{source vector} from which the subspace
activity is harvested, $h^{trg}$ is the \emph{target vector} into which activity
is substituted, and $h^v$ is the resulting intervention vector that we then
use to replace $h^{trg}$ in the model's processing, allowing the model to make
predictions based on a replaced value of variable $\text{v}ar$ following
the intervention.

DAS relies on the notion of counterfactual behavior to create intervention
data to train and evaluate $Q$. For a given SA, we know what the SA's behavior
will be after performing a causal intervention on one of its variables. The
resulting behavior from the SA after intervening on a specific variable and
keeping everything else in the algorithm and task constant is the counterfactual
behavior. This counterfactual behavior can be used as a training signal for
$Q$ using next-token prediction. $Q$ can equivalently learn any
row permutation of the subspaces in $z$, thus we can restrict our searches
to values of $D_{\text{var}}$ that have contiguous non-zero entries. We can
then brute-force search over independent trainings with different values of
$d_{\text{var}}$, selecting the ($Q$,$D_{\text{var}}$) pair with the best
results. Unless otherwise stated, we try values of $d_{\text{var}}$ equal
to either 16 or half of $d_m$ and take the better performing of the two.
See Supplemental Figure~\ref{fig:neuronsweep} for a closer examination
of how $d_{\text{var}}$ affects results.

We perform our causal interventions on individual time steps in the sequence.
We run the model up to an independently sampled timestep $t$ in the target
sequence, taking its latent representation at that point as the target vector,
$h^{trg}_t$. We do the same for the source vector, $h^{src}_u$, at timestep
$u$ from a separate source sequence. We then construct $h^v_t$ using
Equation~\ref{eqn:interchange}, and continue the model's predictions starting
from time $t$, using $h^v_t$ in place of $h^{trg}_t$.

For the LSTM architecture, we perform DAS on a concatenation of the $h$
and $c$ recurrent state vectors \citep{hochreiter1997LSTM}. In the
GRUs, we operate on the recurrent hidden state. In the transformers, we
operate on the residual stream following the first transformer layer
(referred to as the Layer 1 Hidden States in Supplementary
Figure~\ref{fig:tformerarch}) or the input embedding layer. We use
\dasntrain\space intervention samples for training and
\dasnval\space samples for validation and testing. For all data,
we uniformly sample
trial object quantities, and unless otherwise stated, we uniformly
sample intervention time points, $t$ and $u$, from sequence positions
containing demo tokens or response tokens (excluding BOS, trigger, and
EOS tokens). We orthogonalize the rotation matrix
using PyTorch's orthogonal parameterization with default settings. We
train $Q$ with a batch size of \dasbatchsize\space until convergence,
selecting the checkpoint with the best validation performance for
analysis. We use a learning rate of \daslr\space
and an Adam optimizer. See more detail in Supplement~\ref{sup:das}.

\textbf{DAS Evaluation:}
Once our rotation matrix has converged, we can evaluate the quality of
the alignment using the accuracy of the model's predictions on the
counterfactual outputs on held out intervention data. We consider a
trial correct when all deterministic tokens are predicted correctly
using the argmax over logits. We report the proportion of trials
correct as the Interchange Intervention Accuracy (IIA) (as used in
previous work \citep{geiger2023boundless}).

\textbf{DAS Alignment Functions:}
In an effort to understand the solutions employed by the \sameobject\space
RNNs, we introduce relaxations of the orthogonal rotation matrix used in DAS.
We do this by substituting the orthogonal matrix $Q$ with
a general invertible function $f(h)$. We name this class of invertible DAS
functions \emph{alignment functions}
due to their potential to "align" or encode the relationship between
the neural activity and the specified interpretable variables. Formally,
we can write the model's aligned representation, $z$, in
terms of an invertible function, $f$, where $z = f(h)$. In this work, we only
extend DAS to linear cases of $f$ of the form
$f(h) = X(h + b)$ where $X \in R^{d_{m} \times d_{m}}$ is an invertible
symmetric matrix and $b \in R^{d_{m}}$ is a bias vector. Using
$\phi \in \{trg, src\}$ to denote that the same alignment function is
applied to both the target and source vectors before the intervention,
we reformulate Equation~\ref{eqn:interchange} in terms of $f$:
\begin{eqnarray}\label{eq:relation}
    z^{\phi} &=& f(h^{\phi}) = X(h^{\phi} + b) \\
    h^v_t &=& X^{-1}( (1-D_{\text{var}})z^{trg}_t + D_{\text{var}}z^{src}_u ) - b \label{eq:intrvrelation}
\end{eqnarray}
With this formulation, we are able to train $X$ and $b$ using the same
counterfactual sequences used to train $Q$ in Equation~\ref{eqn:interchange}.
We refer to the original DAS analyses as using an
\textbf{Orthogonal Alignment Function} (OAF) and the linear formulation from
Equations~\ref{eq:relation} and \ref{eq:intrvrelation} as the
\textbf{Linear Alignment Function} (LAF). In our experiments, we construct
$X=(MM^\top + \epsilon I)S$ where
$M\in R^{d_m\times d_m}$ is a matrix of learned parameters initially sampled
from a centered gaussian distribution with a standard deviation of $\frac{1}{d_m}$,
$I\in R^{d_m\times d_m}$ is the identity matrix, $\epsilon=0.1$ to prevent
singular values equal to 0, and
$S\in R^{d_m \times d_m}$ is a diagonal matrix to learn a sign
for each column of $X$ using diagonal values
$s_{i,i} = \text{Tanh}(a_i)  + \epsilon (\text{sign}(\text{Tanh}(a_i)))$
where each $a_i$ is a learned parameter and $\epsilon=0.1$ to prevent 0
values.


\subsubsection{Activation Substitutions}\label{sec:methodsactvs}
\textbf{RNN Individual Activation Substitutions:} We explore direct substitutions
of individual neurons in the \multiobject\space trained RNN
models to demonstrate the relative ineffectiveness of individual neuron patching
compared to the rotated subspace patches used in DAS. In these experiments,
we attempt to transfer the \varcount\space in accordance with the
\countupdown\space program by directly replacing the activation value of a
single neuron (A.K.A. dimension) in the RNN's hidden state vector at time step
$t$ with the activation value of the same neuron at time step $u$ from a different
sequence. This is equivalent to Equation~\ref{eqn:interchange} using an identity
rotation with a single, non-zero value in D corresponding to the index of the
specified neuron. We perform these interventions individually for every model
neuron and evaluate the model's IIA using the expected behavior from the
\varcount\space interventions.

\textbf{Transformer Hidden State Substitutions:} A sufficient experiment to
determine whether a Transformer is using Markovian states is to examine its
behavior after replacing all activations in its most recent hidden state
vector from time $t$ from a target sequence with a hidden state vector from time
$u$ from a different source sequence. If the post-intervention behavior matches
that of the source sequence after time $u$, then the state has encoded all
behaviorally relevant information in its activation vector and we can conclude that
the intervened transformer state is Markovian. If the post-intervention behavior
ignores the substitution and matches the target sequence after time $t$, then we
can conclude that the states are anti-Markovian. In the two-layer transformers
used in this work, we only need to perform this intervention on
the output (A.K.A residual stream or hidden states) of the first transformer
layer, as the output of the second layer
can no longer transmit information between token
positions (see Supplement~\ref{sup:sufficiency} for more details). See
Supplement~\ref{sup:ctxintrvdata} for specific intervention data examples.
\begin{figure}[h!]
    \centering
    \includegraphics[width=\textwidth]{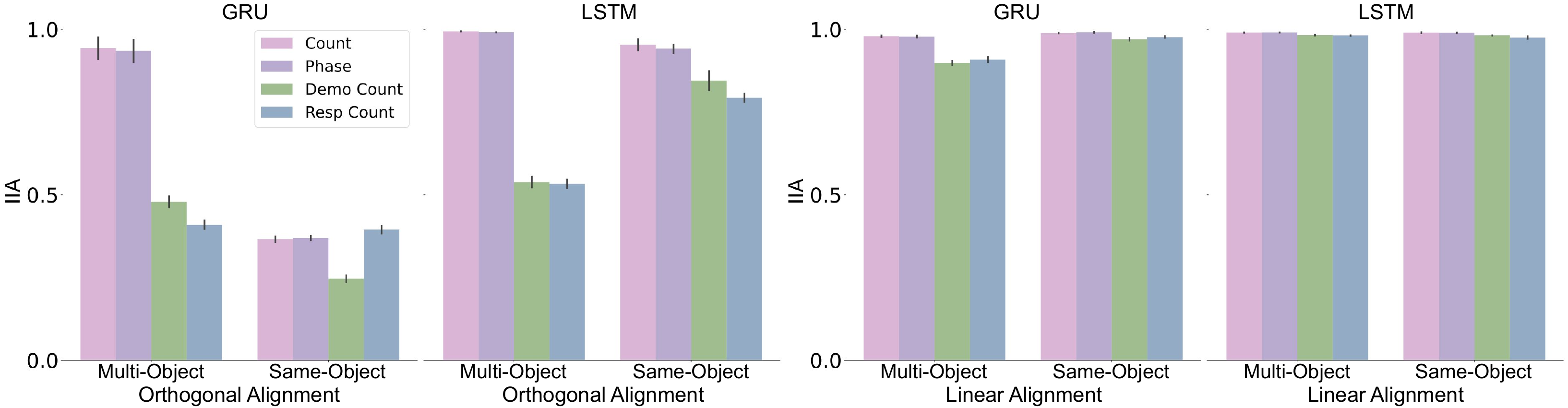}
    \caption{The Interchange intervention accuracy (IIA) for variables
    from different SAs for different tasks and architectures.
    The displayed IIA for the \varcount\space and \varphase\space variables
    comes from the \countupdown\space program. The IIAs for the
    \vardemocount\space and \varrespcount\space variables come from the
    \countupup\space program. IIA measurements show the proportion of
    trials where the model correctly predicts all counterfactual R and
    EOS tokens following a causal intervention. The DAS alignment function
    is displayed below each panel.
    }
    \label{fig:maindasrnns}
\end{figure}
\section{Results}
\subsection{Recurrent Neural Networks}\label{sec:resultsrnns}
\subsubsection{Numeric Neural Variables}\label{sec:resultsdas}
The left side of Figure~\ref{fig:maindasrnns} shows the DAS alignments
using the orthogonal alignment function for RNNs trained on the
\multiobject\space and \sameobject\space tasks.
In the \multiobject\space recurrent models, we see that the most aligned SA
is the \countupdown\space program from the higher IIA in the \varcount\space
and \varphase\space variables compared to the \vardemocount\space and
\varrespcount\space variables from the \countupup\space program.
We use this as supporting evidence for the claim that the
\multiobject\space GRUs and LSTMs possess neural variables for the
\varcount\space and \varphase\space of the task and use a solution
that is causally consistent with the \countupdown\space program.
A visualization of the first two principal components of the solution
found by a \multiobject\space GRU in Supplemental Figure~\ref{fig:multiobjpca}
further supports this interpretation. During the demonstration phase the
projected states take steps in one direction along a line; at the trigger
token, the trajectory jumps in a direction perpendicular to this line;
and during the response phase the states travel parallel to the
demonstration phase trajectory, albeit in the reverse direction.

The existence of these numeric neural variables stands as a
proof of principle that neural systems do not require explicit exposure
to discrete numeric symbols, nor do they need built-in counting principles
to develop symbol-like representations of numbers. Such variables can emerge
in these architectures simply from learning to perform an exact quantity
matching task.


We now focus on the \sameobject\space task where the
demo tokens and resp tokens share a single token id
within the task. We see from Figure~\ref{fig:maindasrnns} that each of
the \sameobject\space GRU OAF alignments have relatively low IIA.
To aid in understanding how the model solves this task variant, we
show the first two principal components of the hidden states of a
\sameobject\space GRU in Supplemental Figure~\ref{fig:sameobjpca}.
As before, the states progress along a line in one direction during the
demonstration phase, and they jump to a line paralleling the first when
transitioning to the response phase. In this case, however, the jump
returns back to the start of the parallel line and then proceeds along
this line in the same direction as in the demonstration phase;
furthermore, the rate of progress and the stopping point in the response
phase both depend on the object count of the trial.
Thus, the main PCs of this model cannot be simply described in terms of
either the \countupdown\space or \countupup\space SAs. We include an
additional DAS analysis in Supplement~\ref{sup:incrupsa} in which we
introduce a new SA that more closely describes our observations from the
PCA in Figure~\ref{fig:sameobjpca}. These additional alignments,
however, also have a relatively low IIA of 58.1\% using an OAF.

We considered whether the state spaces of the \sameobject\space models
could be aligned to an SA using an LAF instead of the OAF. We see in
Figure~\ref{fig:maindasrnns} that the LAF IIA for the \sameobject\space GRUs
reaches very high accuracy (>95\%) for the \varcount\space and \varphase\space
variables of the \countupdown\space program. 
In accordance with the OAF IIA and what we observed in
Figure~\ref{fig:sameobjpca}, we do not conclude that the model has distinct
\varcount\space and \varphase\space variables in the same way as the
\multiobject\space models.
It does mean, however, that such variables are recoverable from its state
space via an invertible linear mapping, and can be causally intervened upon
after such recovery, before being mapped back into the state space of the
model to produce desired, predictable behavior.  

We also see in Figure~\ref{fig:maindasrnns} that the GRU LAF alignments
reach very high IIA (>95\%) for the \vardemocount\space and
\varrespcount\space variables of the \countupup\space program. It appears
that the flexibility of the LAFs allows the states of networks to be aligned
with conceptually distinct programs. In this case, it is possible to compose
the \varcount\space variable of the \countupdown\space program using the difference
between the \vardemocount\space and \varrespcount\space variables from the
\countupup\space program. This perhaps explains why both LAF alignments result in
high IIA for both the Multi- and Same-Object tasks as shown on the right side
of Figure ~\ref{fig:maindasrnns}. We show in Supplemental
Section~\ref{sup:incrupsa} that LAFs do not achieve high IIA
for \emph{all} possible SAs, where the LAF IIA only reaches 72.2\% and 89.4\%
for the \multiobject\space and \sameobject\space GRUs respectively.

In sum, LAFs can allow for causal interventions of high accuracy under a
wider range of circumstances than OAFs, though, this measure alone is
perhaps less diagnostic of the specific SA that best aligns with the
NN's learned solution. We return to LAFs in
Sections~\ref{sec:resultsrelatefxns} and ~\ref{sec:resultsalignvis} to better
explore their theoretic and empirical properties.

\subsubsection{Individual Activation Substitutions}\label{sec:resultsindyactvs}
We performed direct substitutions of individual activation
values in recurrent models' hidden state vectors to demonstrate the importance of
operating on a subspace of the neural population rather than on individual
neurons. We turn our attention to the
raw activation traces in the topmost panel (b) of Figure~\ref{fig:activity},
and note that neurons 12 and 18 (shown in blue and black) have a high correlation
with the \varcount\space of the sequence. These traces came from an LSTM with
$d_m = 20$. In this model, we attempted interchange interventions that transferred
the raw activity from both neurons 12 and 18 in an attempt to transfer the value
of the \varcount. These interventions achieved an IIA of 0.399 on the behavior
generated from the \countupdown\space program. Furthermore,
we observed no consistent pattern of behavior (i.e. off by one errors) following
the interventions. We include this result as a cautionary demonstration
that interpreting correlations without interventions can be problematic;
and the interventions appear to suggest that interpreting raw NN activations
can be misleading.
\begin{figure}[h!]
    \centering
    \includegraphics[width=\textwidth]{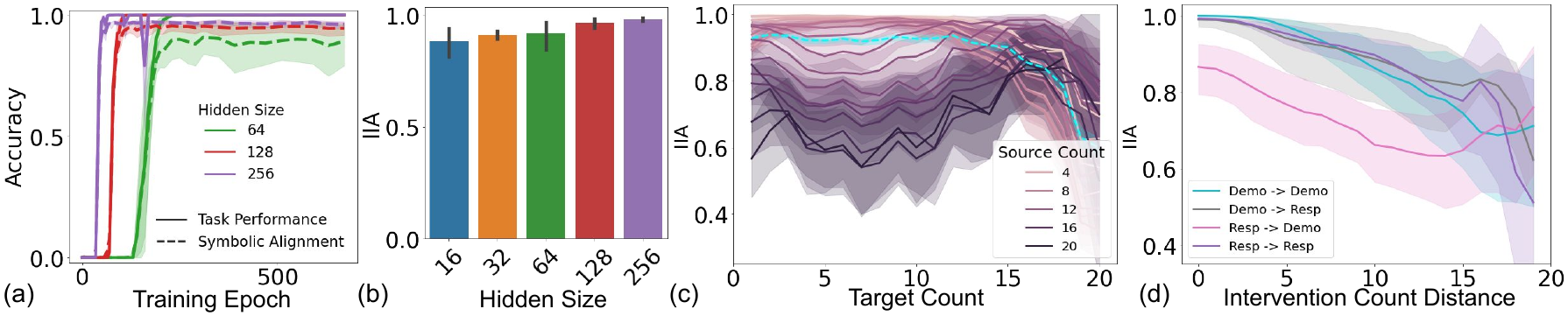}
    \caption{
    In all panels, the IIA comes from DAS using an Orthogonal Alignment trained on
    the \varcount\space variable in the \countupdown\space program. The models are
    all \multiobject\space GRUs.
    (a) Shows task accuracy and IIA over the course of training for architectures
    with different sizes of the recurrent state $h$. We note the correlation between
    IIA and accuracy, with relatively little change as training continues.
    (b) Shows the final IIA for the GRUs as a function of increasing hidden state sizes.
    (c) Shows the IIA from the \dmodel d GRU as a function of the \varcount\space value
    from the source $h^{src}_u$ (denoted by color) and the \varcount\space before
    the intervention in the target $h^{trg}_t$ (shown on the x-axis).
    The cyan, dashed line represents the mean IIA over all interventions for
    a given target count---highlighting the unequal distribution over target
    source pairs.
    (d) DAS IIA from the \dmodel d GRU as a function of the absolute difference between
    the target and source counts. The colors indicate the \varphase\space of $h^{src}$
    on the left and $h^{trg}$ on the right. Panels (c) and (d) show that the value
    of the variable during the interventions somewhat smoothly affect the resulting IIA.
    See Supplemental Section~\ref{sup:dasevaldata} for detail on the data used in panels
    (c) and (d).
    }\label{fig:sizegradience}
\end{figure}
\subsubsection{Graded Neural Variables}\label{sec:resultsgradience}
By increasing the granularity of our analyses, we uncovered a graded
effect of the content of the values involved in the interchange interventions.
We can see this in Figure~\ref{fig:sizegradience} (c) and (d). We see a
gradience in the IIA, where the interventions have a relatively
smooth decrease in IIA when the quantities involved in the intervention are
large and when the intervention quantities have a greater absolute difference.
This indicates that the neural variables have some level of graded continuity
despite the task using fully discrete numeric values.
We refer to such neural variables as \textit{graded} or \textit{symbol-like}.
This notion of graded, symbol-like variables can be contrasted against
cases where the errors are uniformly distributed, independently of the values
used in the interventions.
It is notable that human behavior in many numeric tasks exhibits similar
graded effects, called the size and the distance effect
\citep{Moyer1967numericdistance}. Some researchers note that differential
experience may account for these effects (and showed this in neural
network simulations) \citep{verguts2005sizeeffectmodeling}. In our
experiments, the task training data provides more
experience with smaller numbers, as the models necessarily interact with
smaller quantities every time they interact with larger quantities. This is
perhaps a causal factor for the greater intervention error at larger
numbers, but we do not explore this further. The DAS training data suffers
from a similar issue due to the fact that we use a uniform sampling
procedure for the object quantities that define the
training sequences and we uniformly sample the intervention
indices from appropriate tokens in these sequences. This results in a
disproportionately large number of training interventions containing
smaller values.

The graded neural variables raise the question of how best to interpret
neural networks. We remind ourselves that the ANN is built on a symbolic
computer program, and thus, this program will always align perfectly with
the ANN by definition. The non-trivial goal of our work is to find SAs
that characterize the computations such programs learn such that we can
predict and control the NNs' activity in unified, interpretable ways.
The symbolic gradience that we observe in our models serves as
partial motivation for the LAFs examined in
Section~\ref{sec:resultsrelatefxns}.

\subsubsection{Model Width and Developmental Trajectories}\label{sec:resultsrnnextra}
\textbf{Model Width:} We see in Figure~\ref{fig:sizegradience}(a) that
although many model widths can
solve the \multiobject\space task, increasing the number of
dimensions in the hidden states of the GRUs seems to improve the IIA of the
\countupdown\space alignment. We can also see from
Figure~\ref{fig:sizegradience} (b) that the larger models tend to have better
IIA. Although our results are for RNNs on linear tasks,
an interesting related phenomenon in the LLM literature is the effect
of model scale on performance \citep{brown2020gpt3,kaplan2020scaling}.
We do not concretely explore why increasing dimensionality improves IIA,
but we speculate that with greater dimensionality comes a greater likelihood
that any two variable subspaces will be orthogonal to one another.

\textbf{Developmental Trajectories:}
Turning our attention to the learning trajectories
in Figure~\ref{fig:sizegradience}, we can see that the models'
task accuracy and IIA begin to transition
away from 0\% at similar epochs and plateau at similar epochs. This finding can be
contrasted with an alternative result in which the alignment curves significantly
lag behind the task performance of the models. Alternatively, there could have
been a stronger upward slope of the IIA following the initial performance jump
and plateau. In these hypothetical cases, a possible interpretation
could have been that the network first develops more complex solutions, or
it could have developed unique solutions for many
different input-output pairs and subsequently unified them with further training.
The pattern we observe instead is consistent with the idea that the networks
are biased towards simple, unified strategies early in training.
Perhaps our result is expected from works like \citet{saxe2019} and
\citet{saxe2022race} which show an inherent tendency for NNs trained via
gradient descent to find solutions that share network pathways. This would
provide a driving force towards the demo and resp phases sharing the same
representation of a \varcount\space variable.

\begin{figure}[b!]
    \centering
    \includegraphics[width=\textwidth]{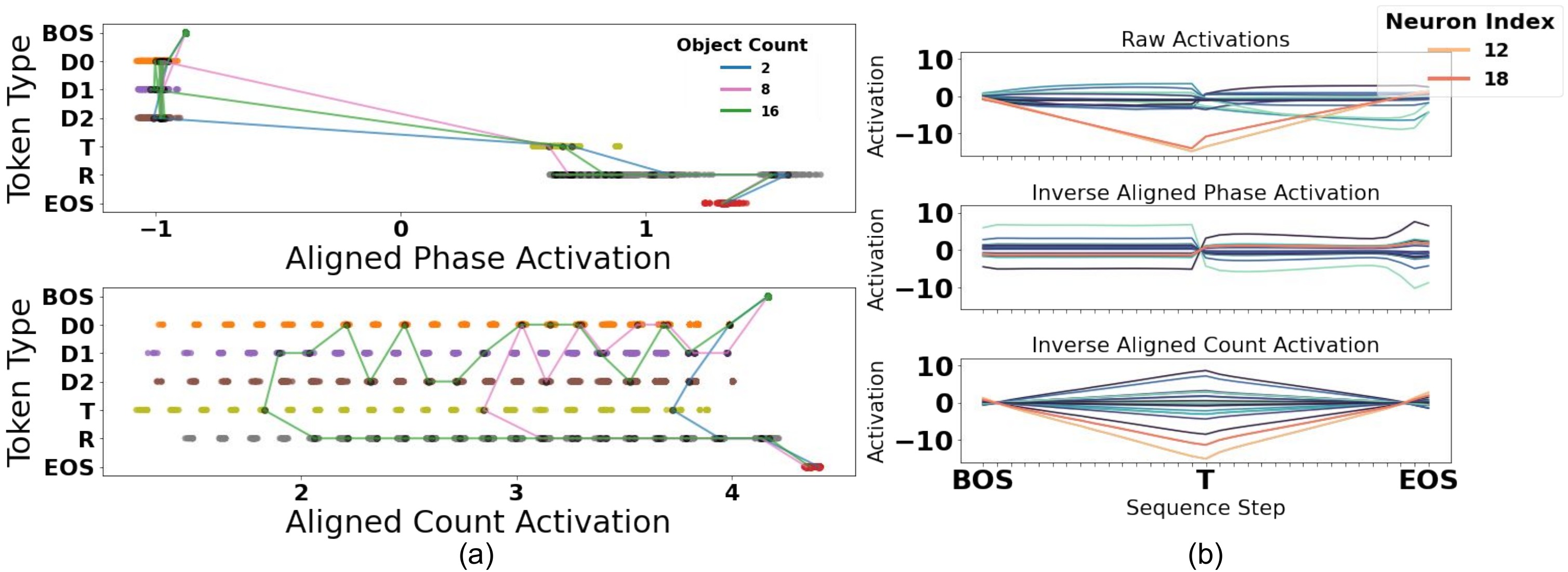}
    \caption{(a) The top panel shows values of $h$ for different trials and
    input token types projected onto the aligned dimension of a 1D linear
    alignment, equal to $DX(h+b)$, for the \varphase\space variable.
    The bottom panel shows the same for the \varcount\space variable. The $h$
    vectors are collected from
    15 trials for each object count ranging from 1-\maxcount\space from a single
    \multiobject\space LSTM of size 20 (size chosen for relative simplicity of
    raw activity). The IIA for the \varphase\space was
    84.7\% and the \varcount\space was 82.6\%. The connecting lines trace the
    states from individual trials with object counts of 2, 8, and 16. Dot
    colors redundantly encode token type.
    (b) These panels show neural variables in the models'
    original neural space and show the importance of causal analysis.
    Each panel shows the activation for
    individual neurons averaged over 15 trials each with an object count 15.
    The x-axis shows steps in the trials. The topmost panel shows the
    raw values. We label two specific neurons (index 12 and 18
    within the $h \in R^{20}$ vector) that have a visibly high correlation
    with the \varcount\space of the sequence. We show in
    Section~\ref{sec:resultsindyactvs} that these two neurons are insufficient
    to causally transfer the \varcount\space between representations. The
    middle and bottom panels show the \varphase\space and \varcount\space
    neural variables projected back into
    the original neural space by inverting the aligned activity using
    $X^{-1}(DX(h+b)) - b$.
    }
    \label{fig:activity}
\end{figure}
\subsubsection{Understanding Alignment Functions}\label{sec:resultsrelatefxns}
In this section, we ask why Linear Alignment Functions (LAFs) improve IIA?
To answer this, we reformulate the model's neural activity, $h$, in
terms of activity component vectors $u_i\in R^{d_m}$:
$h = X^{-1}z - b = Uz-b = \sum_{i=1}^{d_m} z_iu_i - b$, where $X$
and $b$ make up the LAF, $U=X^{-1}$ for notational ease,
$z$ is a vector composed of interpretable subspaces from
Equation~\ref{eq:zdef}, and $z_i$ refers to the value of the
$i^{\text{th}}$ dimension of $z$.
The interchange intervention in Equation~\ref{eq:intrvrelation} is equivalent
to exchanging weighted activity components $z_i u_i$:
\begin{eqnarray}
    h^v &=& U( D_{\text{var}}z^{src} + (1-D_{\text{var}})z^{trg} ) - b 
    =
      - b + 
      \sum_{i=1}^{d_{m}} \mathbf{1}_{\{i \leq d_{\text{v}ar}\}} z^{src}_iu_i +
      \sum_{i=1}^{d_{m}} \mathbf{1}_{\{i > d_{\text{v}ar}\}} z^{trg}_iu_i
       \\
    h^v &=& 
      - b +
      \sum_{i=1}^{d_{\text{var}}} z^{src}_iu_i +
      \sum_{i=1+d_{\text{var}}}^{d_m} z^{trg}_iu_i =
      - b +
      \sum_{i=1}^{d_{\text{var}}} z^{src}_iu_{\text{var},i} +
      \sum_{i=1+d_{\text{var}}}^{d_m} z^{trg}_iu_{\cancel{\text{var}},i}
      \label{eq:weightedcomps}
\end{eqnarray}
Where $u_{\text{var},i}$ indicates that the activity corresponds to the
intervened variable subspace and $u_{\cancel{\text{var}},i}$ is all other
activity. If $U$ is orthogonal, then by definition, each inner product
$\langle u_i, u_j\rangle=0$ when $i\neq j$. Thus
$\langle z_iu_i, z_ju_j\rangle=z_iz_j\langle u_i, u_j\rangle=0$ too. If
$U$ is orthogonal, then so is its inverse, and thus
$\langle U^{-1}z_iu_i, U^{-1}z_ju_j\rangle=0$ when $i\neq j$ due to
orthogonal matrices preserving inner products.
Thus when using an orthogonal alignment function, the intervened
subspaces are also orthogonal in the original neural space.
This is not necessarily the case, however, for LAFs, where $U$ is
a linear invertible matrix because
$\langle U^{-1}z_iu_i, U^{-1}z_ju_j\rangle$ need not
be equal to 0. This means that the LAF can
allow the intervened subspaces to be non-orthogonal in the original
neural space, thus allowing the LAF to formulate each variable from
the SA in linear terms of the other variables.

We demonstrate this by first writing $h$ as a combination of alignment
vectors corresponding to each variable. Using the \varcount\space
and \varphase\space variables as an example, we decompose $h$ as follows:
\begin{equation}\label{eq:hdecomp}
h = U_{\text{count}}\Vec{z}_{\text{count}} +
     U_{\text{phase}}\Vec{z}_{\text{phase}} +
     U_{extra}\Vec{z}_{extra} - b
\end{equation}
where $\Vec{z}_{\text{v}ar} \in R^{d_{\text{v}ar}}$ and
$U_{\text{v}ar} \in R^{d_m\times d_{\text{var}}}$. We can now solve for $\Vec{z}_{count}$
by rearranging Equation~\ref{eq:hdecomp}:
\begin{equation}\label{eq:zdecomp}
U_{count}\Vec{z}_{count} = h - U_{phase}\Vec{z}_{phase} - U_{extra}\Vec{z}_{extra} + b
\end{equation}
In the case where $U$ is orthogonal, then $U_{count}^{-1} = U_{count}^{\top}$
and $U_{count}^{\top}U_{phase} = U_{count}^{\top}U_{extra} = 0$,
meaning that $\Vec{z}_{count}$ must
be defined independently of $\Vec{z}_{phase}$. This is not the case,
however, for a non-orthogonal $U$, which means that each $\Vec{z}_{var}$ can
be written in linear terms of the other variables in the SA when using LAFs.

\subsubsection{Visualization of symbolic algorithm variables with alignment functions}\label{sec:resultsalignvis}

While IIA alone may not be fully diagnostic of the SA that best characterizes
the computations performed by an NN, further analysis based on a successful
alignment can potentially shed further light on the algorithm a network is
using, and it can aid in understanding how neural variables are encoded in
the raw neural activity.
We provide Figure~\ref{fig:activity} (a) and (b) to illustrate these points.
The Figures were obtained using an LAF with $d_{\text{var}}=1$ aligned to a
\multiobject\space LSTM with a hidden dimensionality of 20 ($d_m = 20$).
We chose $d_{\text{var}}=1$ and $d_m = 20$ for visualization purposes, and
we use an LAF for its improved IIA when using $d_{\text{var}}=1$ relative to
an OAF on the smaller model size.

Figure~\ref{fig:activity}(a) shows a set of $h$ vectors projected into the
1-dimensional, aligned \varphase\space and \varcount\space subspaces taken from
the LSTM at many different time steps and trials; we illustrate trajectories in
each subspace for individual trials with object counts of 2, 8, and 16. The
vertical axis in each panel is used to facilitate the visualization of individual
trajectories and their relation to specific token types.
One can see in the top panel of (a) how the \varphase\space variable has a
value of -1 with slight jitter independent of the demonstration token instance.
The \varphase\space then transitions to a positive value at the T token, and
stays positive throughout the response phase, though, with some contamination by
the \varcount. Similarly, one can see in the bottom panel of
~\ref{fig:activity}(a) how the \varcount\space variable
progresses from right to left in integer-like steps during the demonstration
phase, takes one step further left at the T token, then jumps back 2 steps to the
right at the first R token to correctly represent the decremented value of the
\varcount\space after the first response token. Lastly, it steps rightward
until the \varcount\space reaches 0, after which the EOS is produced.

Figure~\ref{fig:activity}(b) shows the raw neural activity in the top panel for
all 20 of the neurons in the model averaged over 15 sampled trials with object
counts of 15,
and the bottom two panels show the inverse of \emph{only} the aligned activity
for each individual variable, equal to $z_{\text{phase}}u_{\text{phase}}-b$
in the middle panel and $z_{\text{count}}u_{\text{count}}-b$ in the bottom.
We provide this figure as a demonstration of how to view the causally relevant
neural activity in the NN's original neural space through the
lens of the aligned variables. We can see that many of the raw
neurons can play a causal role in both the \varphase\space and \varcount\space
neural coding, confirming notions of distributed coding from prior work
\citep{PDP1,Smolensky1988,elhage2022superposition,olah2023compositionsuperposition}.

\subsection{Transformers}\label{sec:resultstformers}
In this section, we demonstrate through empirical and theoretical means that
shallow transformers solve the task by recomputing the solution to the task at
each step in the sequence. We refer to this class of solutions as anti-Markovian,
named for their inductive bias against cumulative, Markovian states. We
begin by demonstrating that using the previous transformer layer's outputs as
the inputs to the attention mechanism in the next layer restricts transformers
from using
Markovian solutions that use more steps than attention layers. We then
demonstrate a theoretical solution
for simplified versions of the \singleobject\space and \multiobject\space numeric
equivalence tasks in one layer NoPE transformer architectures, and we causally
verify that such a solution emerges empirically. Lastly, we show through causal
interventions that similar solutions can emerge in two layer RoPE transformers.

\subsubsection{Anti-Markovian States}\label{sec:resultsantimarkov}
In this section, we demonstrate why Transformer solutions that use
Markovian states in the residual stream require a new attention layer
for every new step in the sequence. To show this, we focus on a simplified
transformer architecture that only includes an embedding layer and the self-attention
mechanism within each layer. To justify this simplification, we note that
the attention mechanism is the only mechanism in the transformer that provides an
opportunity to transmit state information between token positions in the sequence.
With this simplification, we can write the output of a single transformer layer as:
\begin{equation}
\begin{bmatrix}
    h_0^\ell & h_1^\ell  & ... & h_t^\ell
\end{bmatrix}
= 
\begin{bmatrix}
    h_0^{\ell-1} & h_1^{\ell-1}  & ... & h_t^{\ell-1}
\end{bmatrix} + \text{attn}_\ell(
\begin{bmatrix}
    h_0^{\ell-1} & h_1^{\ell-1}  & ... & h_t^{\ell-1}
\end{bmatrix}
)
\end{equation}
where $h_t^\ell\in R^d$ are column vectors from the transformer residual stream,
$\ell$ denotes the attention layer where $\ell=0$ is the output of the embedding layer,
$t$ refers to the positional index in the sequence, and $\text{attn}_\ell(x)$ refers
to the attention mechanism. We denote a cumulative state at step $m$ in a Markov
chain as $s_m$, and we denote the encoded state in a residual stream vector as
$h^{\ell, (s_m)}_t$. We assume that the attn function can only produce and
encode $s_{m+1}$ at time $t$ if $s_m$ is already encoded at time $<t$, and we
assume that $s_0$ is produced in the embedding layer. Then the
cumulative state gets updated with each transformer layer as follows:
\begin{eqnarray*}
\begin{bmatrix}
    h_0^{0,(s_0)} & h_1^0  & h_2^0 & ... & h_t^0
\end{bmatrix}
\!\!\!\!&=&\!\!\!\!
\text{Embedding}( x_0, x_1, x_2, ...,  x_t ) \\
\begin{bmatrix}
    {h}_0^{1,(s_0)} & {h}_1^{1,(s_1)}  & h_2^1 & ... & h_t^1
\end{bmatrix}
\!\!\!\!&=&\!\!\!\!
\begin{bmatrix}
    {h}_0^{0,(s_0)} & h_1^{0}  & h_2^0 & ... & h_t^{0}
\end{bmatrix} + \\
\!\!\!\!&&\!\!\!\!
\text{attn}_1(
\begin{bmatrix}
    {h}_0^{0,(s_0)} & h_1^{0}  & h_2^0 &  ... & h_t^{0}
\end{bmatrix}
) \\
\begin{bmatrix}
    {h}_0^{2,(s_0)} & {h}_1^{2,(s_1)}  & {h}_2^{2,(s_2)}  & ... & h_t^2
\end{bmatrix}
\!\!\!\!&=&\!\!\!\!
\begin{bmatrix}
    {h}_0^{1,(s_0)} & {h}_1^{1,(s_1)} & h_2^1 & ... & h_t^1
\end{bmatrix} + \\
\!\!\!\!&&\!\!\!\!
\text{attn}_2(
\begin{bmatrix}
    {h}_0^{1,(s_0)} & {h}_1^{1,(s_1)} & h_2^1  & ... & h_t^1
\end{bmatrix}
) \\
\begin{bmatrix}
    {h}_0^{t,(s_0)} & {h}_1^{t,(s_1)}  & {h}_2^{t,(s_2)}  & ... & {h}_t^{t,(s_t)}
\end{bmatrix}
\!\!\!\!&=&\!\!\!\!
\begin{bmatrix}
    {h}_0^{t-1,(s_0)} & {h}_1^{t-1,(s_1)} & {h}_2^{t-1,(s_2)}& ... & h_t^{t-1}
\end{bmatrix} +\\
\!\!\!\!&&\!\!\!\!
\text{attn}_t(
\begin{bmatrix}
    {h}_0^{t-1,(s_0)}& {h}_1^{t-1,(s_1)}& {h}_2^{t-1,(s_2)} & ... & h_t^{t-1}
\end{bmatrix}
)
\end{eqnarray*}
Where $x_t$ denotes the input token id at time $t$. We can see that in the
best case scenario, the cumulative state can only be transmitted and updated
one layer at a time (we provide a more formal proof in
Supplement~\ref{sup:antimarkovproof}). Thus the two layer transformers in
our work are architecturally insufficient for using a solution that involves
Markovian states.

We experimentally verify that the two layer RoPE transformers used in
this work use anti-Markovian states by performing Transformer Hidden State
Substitutions as outlined in Methods Section~\ref{sec:methodsactvs}. Indeed,
these substitutions leave the NNs' behavior largely unaffected with an IIA of
\MultiObjTransformerCtxDistrDASMean\space on the original behavior in the
\multiobject\space RoPE transformers and \VaryLenTransformerCtxDistrDASMean\space
for the \varylen\space \multiobject\space RoPE transformers.

We note that generative techniques like scratch pad \citep{nye2021scratchpad}
and Chain-of-Thought (CoT) \citep{wei2023chainofthought} allow for transformers to
track a cumulative state in the form of self-generated input embeddings.
We might expect recurrent models to benefit less from CoT in this respect.

%
%
\subsubsection{Simplified NoPE Transformers}\label{sec:resultsNoPE}
To better understand how a transformer could impement an anti-Markovian
solution to the \multiobject\space task,
we include a theoretical treatment of a single-layer NoPE Transformer that is
trained on a simplified version of the \multiobject\space task: it excludes
the BOS and T tokens from the sequences and only has one possible
demo token instance D that is different from the resp token instance R
(i.e. an object count of 2 would result in the sequence "D D R R E").
The self-attention calculation for a single query $q_r \in R^d$
from a response token, denoted by the subscript $r$, is as follows:
\begin{equation}
    \text{Attention}(q_r, K, V) = V\big(\text{softmax}(\frac{K^\top q_r}{\sqrt{d}})\big) =
    \sum_{i=1}^n{ \frac{ e^{\frac{q_r^\top k_i}{\sqrt{d}}} }{\sum_{j=1}^n e^{\frac{q_r^\top k_j}{\sqrt{d}}}} v_i } =
    \sum_{i=1}^n{ \frac{ s^{r}_i }{\sum_{j=1}^n s^{r}_j} v_i } =
    \frac{ 1 }{\sum_{j=1}^n s^{r}_j} \sum_{i=1}^n{ s^{r}_i v_i }
\end{equation}
Where $d$ is the dimensionality of the model, $n$ is the sequence length,
$K \in R^{d\times n}$ is a matrix of column vector keys, $V \in R^{d\times n}$
is a matrix of column vectors $v_i$, and $s^r_i = e^{\frac{q_r^\top k_i}{\sqrt{d}}}$,
using $i$ to denote the positional index of the key and the superscript $r$
to denote that the $q$ came from a response token. We refer to $s^r_i v_i$
as the strength-value of the $i^{\text{th}}$ token for the query $q_r$.

In the first layer following the embeddings in a NoPE transformer, each of the queries
for the response tokens will
produce equal strength-values for a given key-value pair regardless of the position
from which the response token and demo tokens originated. This is because NoPE does
not add positional information to the embeddings. Thus, assuming that the attention
mechanism is performing a sum of the count contributions from each token in the
sequence, we should be able to use the $s^r_i v_i$ to increment and decrement the
model's decision to produce the EOS token from any given response token in the
following way:
\begin{equation}\label{eqn:increment}
    \text{IncrementedAttention}(q_r, K, V) =
    \frac{1}{s^r_r + \sum_{j=1}^n s^r_j}\big( s_r^r v_r + \sum_{i=1}^n{ s_i v_i } \big)
\end{equation}
Where the subscript $r$ in the strength $s_r$ and value $v_r$ denotes that the
originating token for the key-value pair is a response token. We can
decrement the count using a key-value pair from a demonstration token.
To verify our theoretical treatment, we performed a simulation using a
single-layer NoPE
transformer trained on the simplified \singleobject\space task. Using the
strength-value additions outlined in Equation~\ref{eqn:increment},
we were able to change the position at which the
transformer produced the EOS token with 100\% accuracy. We include results
for other transformer architecture variants in
Supplemental Figure~\ref{fig:maindastformers} (c).

\subsubsection{RoPE Transformers}\label{sec:resultsRoPE}
To determine how the RoPE transformers perform the tasks, we first looked at the
attention weights for both of its two layers (see Supplemental
Figure~\ref{fig:attnrope}). The R and EOS queries give surprisingly little
attention to the R tokens. In Supplemental Figure~\ref{fig:maindastformers},
we show DAS results on the \varlastval\space variable from the \distrsoln\space
SA where a numeric value is assigned to each token and the values of all previous
tokens are summed at each step in the sequence. The \multiobject\space transformers
achieved an IIA of 0.800 for this alignment. We took this to mean that the
\multiobject\space transformers were at least partially using a positional
readout to solve the task. To understand transformer solutions that rely less
on a positional readout, we also examined a set of transformers trained on the
\varylen\space variant of the \multiobject\space task that disrupts
count-position correlations. These transformers achieved a higher IIA of
0.935 for the same DAS alignment. The lower IIA of the original
\multiobject\space transformers is consistent with the notion that they rely,
in part, on a positional readout, rather than a summing operation, to solve the
task.
\section{Conclusion}
In this work we used both causal and correlational methods to interpret
emergent representations of numbers in several types of NNs.
We used these methods to discover and characterize graded, symbol-like
number variables within the representations of two types of RNNs; we
introduced an extension of DAS that adds flexibility our ability characterize
neural activity in terms of functions of interpretable symbolic variables;
we explored theoretical and empirical transformer
solutions to the tasks; and we showed that transformers must use
anti-Markovian solutions in the absence of sufficient layers. 

As we have noted before in this article, ANNs are by definition symbolic
programs, but the algorithms they learn are latent in their
connection weights and are not directly observable by inspection of the
ANN code. Our goal in aligning NNs and SAs is to find predictive,
interpretable, and controllable ways of understanding the learned neural
activity. Because NNs are example-driven, gradient-based learners, their
learned solutions do not necessarily need to match interpretable SAs
exactly. Under idealized conditions, they can come to conform more and more
precisely to such algorithms, but they generally do so gradually through
training. We believe these points characterize many aspects of the algorithms
humans and animals learn through experience, making NNs good models of many
aspects of human cognition and development.

When networks learn to conform to simple SAs, they tend to do so in narrow
task settings, disconnected from the richness and complexity of real experience
\cite{servan1991graded}. This situation applies to our models, whose
experience is extremely narrow, limited to correct instances of a
sequence-based numeric equivalence task. In more complex cases, such as natural
language, debate has raged for decades about whether NNs are adequate for
extracting the symbolic rules that many argue underlie human language
abilities. In our view \cite{mcclelland2015capturing}, such rules are often
useful characterizations for the goals of interpretability, but should not be
embraced as the complete story without skepticism. Nevertheless, we strongly
encourage work that seeks to understand the computations of neural systems
through their alignment with SAs.
%
\section{Acknowledgments}
Thank you to the PDP Lab for thoughtful discussion and the
Stanford Psychology department for funding. Thank you to
the ICLR and TMLR reviewers for greatly improving the paper.
And huge thank you to Zen Wu for great discussions and
assisting with numerous questions about DAS early in the
project's development.

\bibliography{tmlr}
\bibliographystyle{tmlr}


\pagebreak
\appendix
\section{Appendix / supplemental material}
\subsection{Additional Figures}
\begin{figure}[H]
    \centering
    \includegraphics[width=0.5\textwidth]{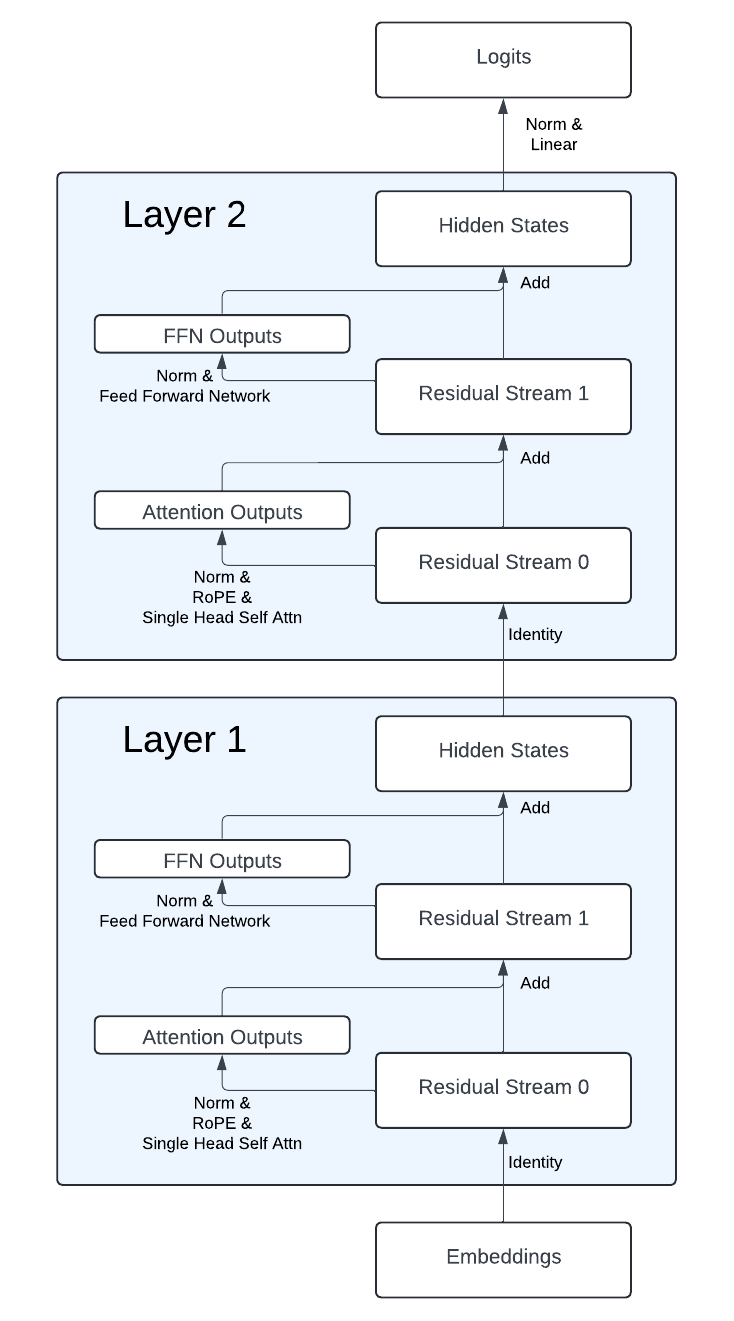}
    \caption{
    Diagram of the main transformer architecture used in this work. The
    white rectangles represent activation vectors. The arrows represent
    model operations. All normalizations are Layer
    Norms \citep{ba2016layernorm}. The majority of the DAS interchange
    interventions are performed on Hidden State activation vectors from
    Layer 1 at individual time-steps. We offer further granularity in
    the \varlastval\space interventions by performing DAS on the embeddings
    that are projected into the key and value vectors for the Layer 1
    self-attention.
    }
    \label{fig:tformerarch}
\end{figure}
\begin{table}[!]
\caption{
The DAS results for each model and task variant. Each alignment function
was trained on a single causal variable with a $d_{\text{var}}$
(subspace size) of the better performing out of 16 or 64 dimensions.
Performance values are reported as IIA under the task name.
}
\label{tab:dasrnns}
\begin{center}
\begin{tabular}{c c c c c c c}
\multicolumn{1}{c}{\bf Model} & \multicolumn{1}{c}{\bf Alignment} & \multicolumn{1}{c}{\bf Algorithm} & \multicolumn{1}{c}{\bf Variable} & \multicolumn{1}{c}{\bf Multi-Object} & \multicolumn{1}{c}{\bf Same-Object} & \multicolumn{1}{c}{\bf Single-Object}  \\
\\ \hline
 GRU & Orthogonal & Up, Down & Count & 0.9464 & 0.3698 & 0.908 \\
\hline
 GRU & Orthogonal & Up, Down & Phase & 0.9368 & 0.373 & 0.889 \\
\hline
 GRU & Orthogonal & Up, Up & Demo Count & 0.4807 & 0.2503 & 0.606 \\
\hline
 GRU & Orthogonal & Up, Up & Resp Count & 0.4173 & 0.4024 & 0.477 \\
\hline
 GRU & Linear & Up, Down & Count & 0.9906 & 0.991 & 0.995 \\
\hline
 GRU & Linear & Up, Down & Phase & 0.9884 & 0.9922 & 0.991 \\
\hline
 GRU & Linear & Up, Up & Demo Count & 0.9174 & 0.9744 & 0.984 \\
\hline
 GRU & Linear & Up, Up & Resp Count & 0.9223 & 0.9808 & 0.989 \\
\hline
 LSTM & Orthogonal & Up, Down & Count & 0.993 & 0.958 & 0.989 \\
\hline
 LSTM & Orthogonal & Up, Down & Phase & 0.991 & 0.95 & 0.991 \\
\hline
 LSTM & Orthogonal & Up, Up & Demo Count & 0.5416 & 0.86 & 0.409 \\
\hline
 LSTM & Orthogonal & Up, Up & Resp Count & 0.5374 & 0.8007 & 0.439 \\
\hline
 LSTM & Linear & Up, Down & Count & 0.9928 & 0.9922 & 0.992 \\
\hline
 LSTM & Linear & Up, Down & Phase & 0.9914 & 0.9912 & 0.99 \\
\hline
 LSTM & Linear & Up, Up & Demo Count & 0.9846 & 0.9846 & 0.990 \\
\hline
 LSTM & Linear & Up, Up & Resp Count & 0.9862 & 0.9697 & 0.991 \\
\end{tabular}
\end{center}
\end{table}
\begin{table}[!]
\caption{
The DAS results for each model and task variant using a linear alignment
function with a $d_{\text{var}}=1$ (subspace size). Each alignment function
was trained on a single causal variable.
Performance values are reported as IIA under the task name.
}
\label{tab:1dlineardasrnns}
\begin{center}
\begin{tabular}{c c c c c c c}
\multicolumn{1}{c}{\bf Model} & \multicolumn{1}{c}{\bf Alignment} & \multicolumn{1}{c}{\bf Algorithm} & \multicolumn{1}{c}{\bf Variable} &\multicolumn{1}{c}{\bf $d_{\text{var}}$} &  \multicolumn{1}{c}{\bf Multi-Object} & \multicolumn{1}{c}{\bf Same-Object} \\
\\ \hline
 GRU & Linear & Count Up, Count Down & Count & 1 & 0.9436 & 0.68 \\
\hline   
 GRU & Linear & Count Up, Count Down & Phase & 1 & 0.8568 & 0.870 \\
\hline 
 GRU & Linear & Count Up, Count Up & Demo Count & 1 & 0.758 & 0.507\\
\hline
 GRU & Linear & Count Up, Count Up & Resp Count & 1 & 0.8339 & 0.890\\
\hline
 LSTM & Linear & Count Up, Count Down & Count & 1 & 0.9236 & 0.864\\
\hline
 LSTM & Linear & Count Up, Count Down & Phase & 1 & 0.955 & 0.86 \\
\hline
 LSTM & Linear & Count Up, Count Up & Demo Count & 1 & 0.912 & 0.775 \\
\hline
 LSTM & Linear & Count Up, Count Up & Resp Count & 1 & 0.9494 & 0.900 \\
\hline
\end{tabular}
\end{center}
\end{table}
\begin{table}[!]
\caption{
The DAS results for the transformers. Each DAS training was
performed on a single causal variable with a $d_{\text{var}}$
(subspace size) of the better performing out of 24 or 64 dimensions.
Performance values are reported as IIA under the task name.
}
\label{tab:tformerdas}
\begin{center}
\begin{tabular}{c c c c c c}
\multicolumn{1}{c}{\bf Model} &
\multicolumn{1}{c}{\bf Alignment} &
\multicolumn{1}{c}{\bf Algorithm} &
\multicolumn{1}{c}{\bf Variable} &
\multicolumn{1}{c}{\bf Multi-Object} &
\multicolumn{1}{c}{\bf Same-Object} \\
\\ \hline
 NoPE Transformer & Orthogonal & Context Distributed & Input Value & 0.882 & 0.982 \\
\hline
 NoPE Transformer & Orthogonal & Count Up, Count Down & Count & 0.1112 & 0.110 \\
\hline
 RoPE Transformer & Orthogonal & Context Distributed & Input Value & 0.8004 & 0.935 \\
\hline
 RoPE Transformer & Orthogonal & Count Up, Count Down & Count & 0.1274 & 0.124 \\
\hline
\end{tabular}
\end{center}
\end{table}
\begin{figure}[H]
    \centering
    \includegraphics[width=0.85\textwidth]{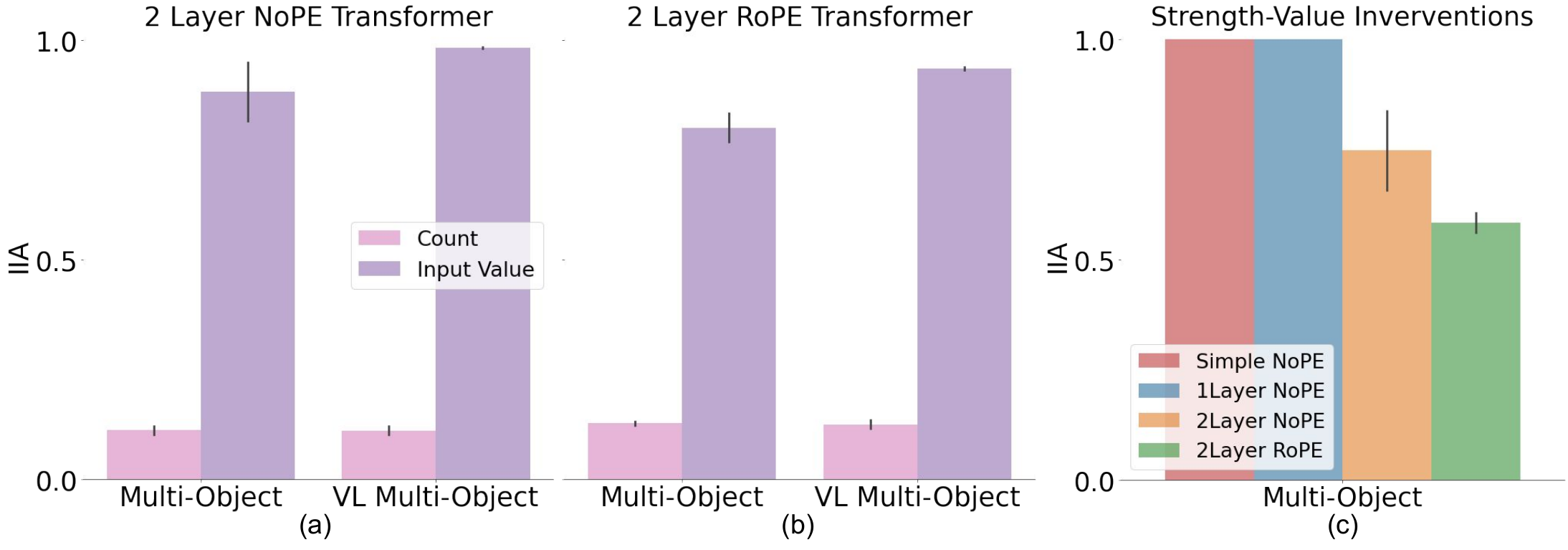}
    \caption{
    (a) and (b) show the interchange intervention accuracy (IIA) on the \varcount\space
    from the \countupdown\space program and the \varlastval\space from the
    \distrsoln\space program aligned to the Transformer architectures using DAS
    with an Orthogonal alignment function. VL denotes models trained on the
    \varylen\space version of the task. The \varlastval\space encodes an assigned
    value (+1, -1, or 0) to each incoming token that is used to recalculate the
    count at each step in the sequence. The DAS analysis is applied to the model
    embeddings for the \varlastval, and the residual stream after the first transformer
    layer for the \varcount. We can see that the \varylen\space transformers have
    stronger alignment to the \varlastval\space variable---consistent with an
    interpretation in which the \multiobject\space transformers can rely, to some degree,
    on positional information.
    (c) IIA for strength-value interventions described in Section~\ref{sec:resultsNoPE}.
    These interventions add and subtract from the count using the strength-value within
    an attention computation. Strength-values are computed from the last response
    query, key, and value in the sequence from the layer in which interventions are
    performed. The displayed IIA is taken from the better performing of the possible
    attention layers.
    }
    \label{fig:maindastformers}
\end{figure}
\begin{figure}[H]
    \centering
    \includegraphics[width=0.97\textwidth]{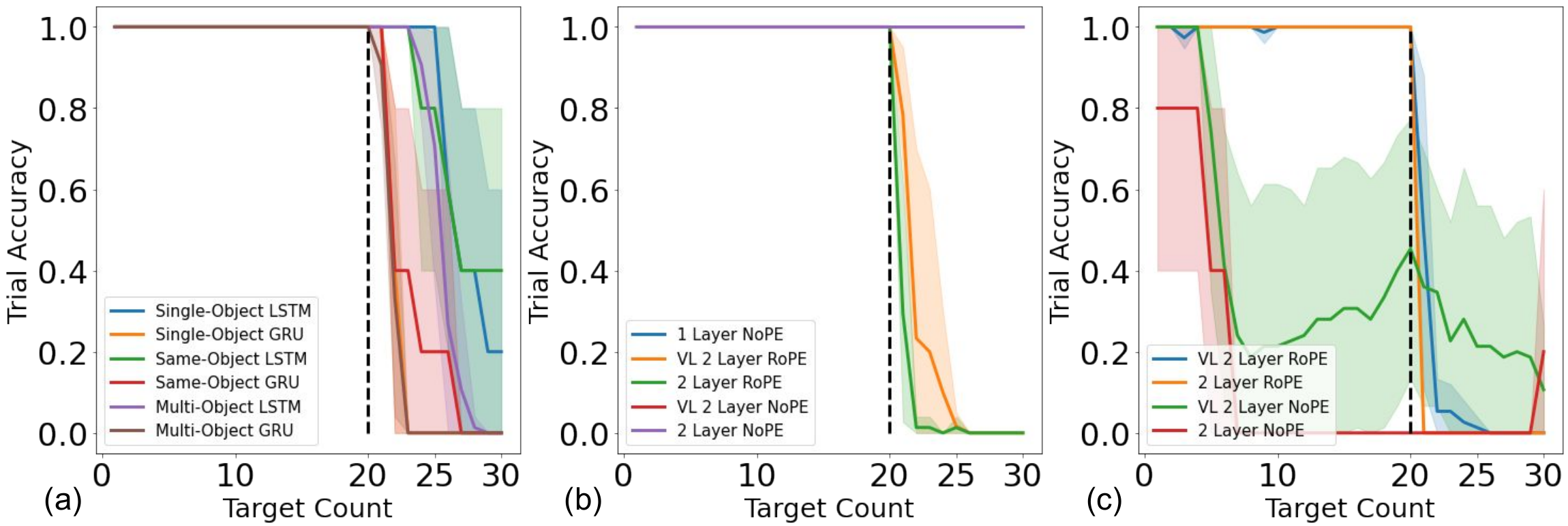}
    \caption{
        (a) RNN task performance measured as the proportion of trials correct.
        Object quantity refers to the number of demo tokens in a sequence
        preceding the trigger token. The evaluation data consists of
        15 sampled sequences (even when only one configuration exists for that object
        quantity).
        (b) Transformer performance on the \multiobject\space task.
        VL indicates the \varylen\space version of the task. One model seed
        was dropped from each the NoPE and RoPE models trained on the
        \varylen\space \multiobject\space task due to lower than 99\% accuracy. 
        (c) Transformer performance on the \sameobject\space task.
        VL indicates the \varylen\space version of the task. All NoPE model
        seeds performed below 99\% accuracy on the \sameobject\space task. 
    }
    \label{fig:taskaccs}
\end{figure}
\begin{figure}[H]
    \centering
    \includegraphics[width=0.7\textwidth]{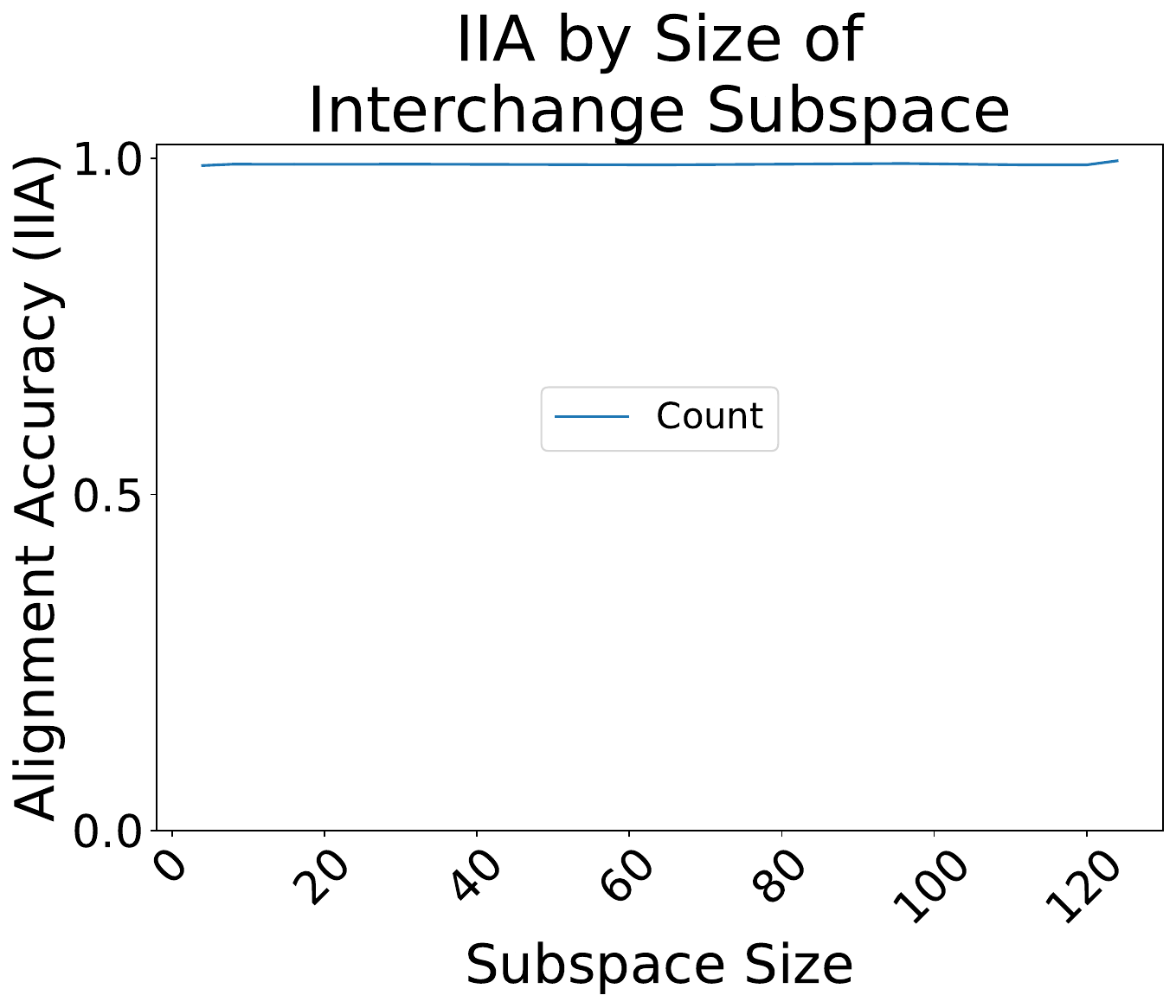}
    \caption{
        An exploration of the performance of the orthogonal DAS alignment
        as a function of the size of the interchange subspace for a randomly
        selected \multiobject\space LSTM model seed on the \varcount\space
        variable.
        The x axis shows $d_{\text{count}}$ while the y axis shows IIA.
        This is the number of dimensions substituted in the intervention.
    }
    \label{fig:neuronsweep}
\end{figure}
\begin{figure}[h]
    \centering
    \includegraphics[width=0.7\textwidth]{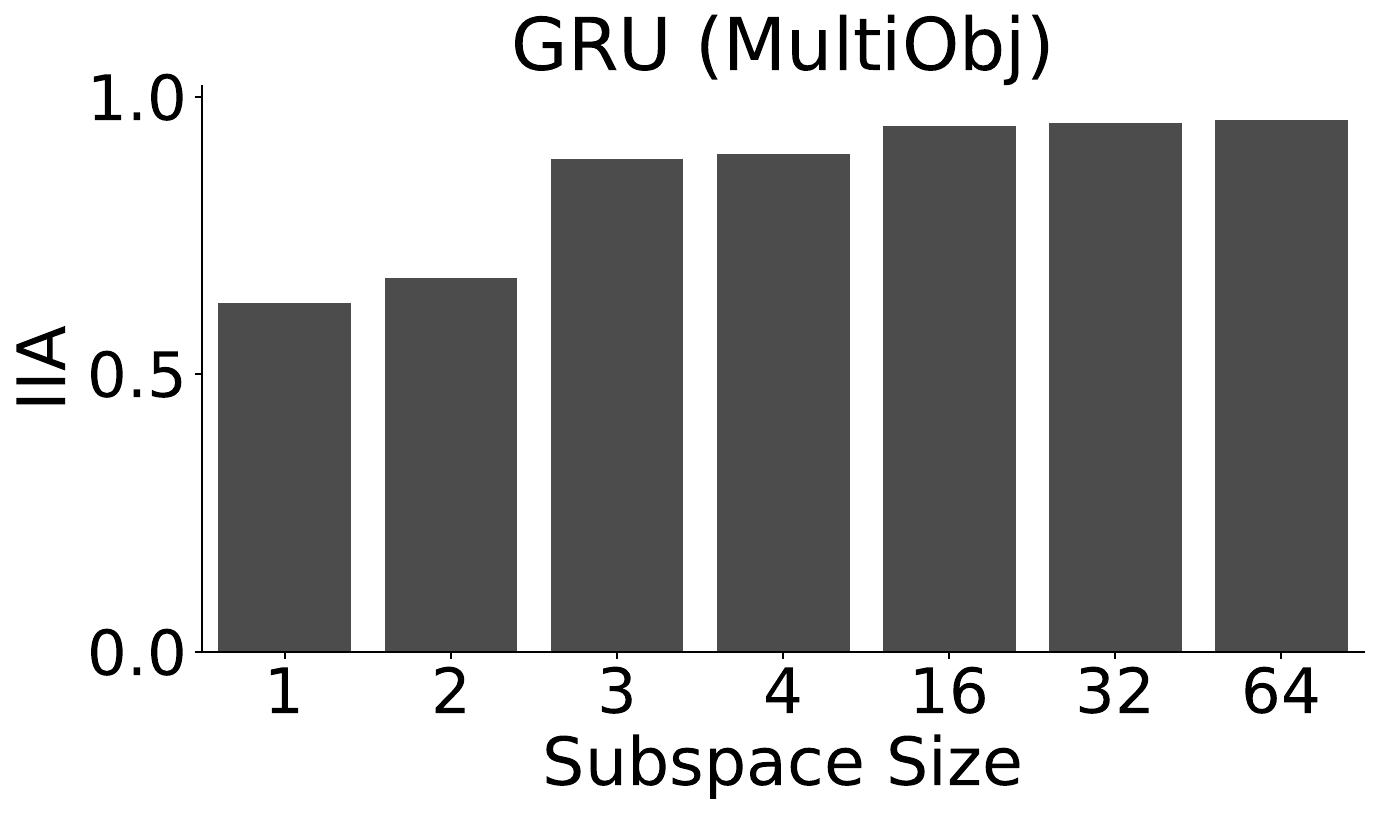}
    \caption{
        An exploration of the performance of the orthogonal DAS alignment
        as a function of the size of the interchange subspace for a randomly
        selected \multiobject\space GRU model seed on the \varcount\space
        variable.
        The x axis shows $d_{\text{count}}$ while the y axis shows IIA.
        This is the number of dimensions substituted in the intervention.
    }
    \label{fig:neuronsweepgru}
\end{figure}
\begin{figure}[H]
    \centering
    \includegraphics[width=0.97\textwidth]{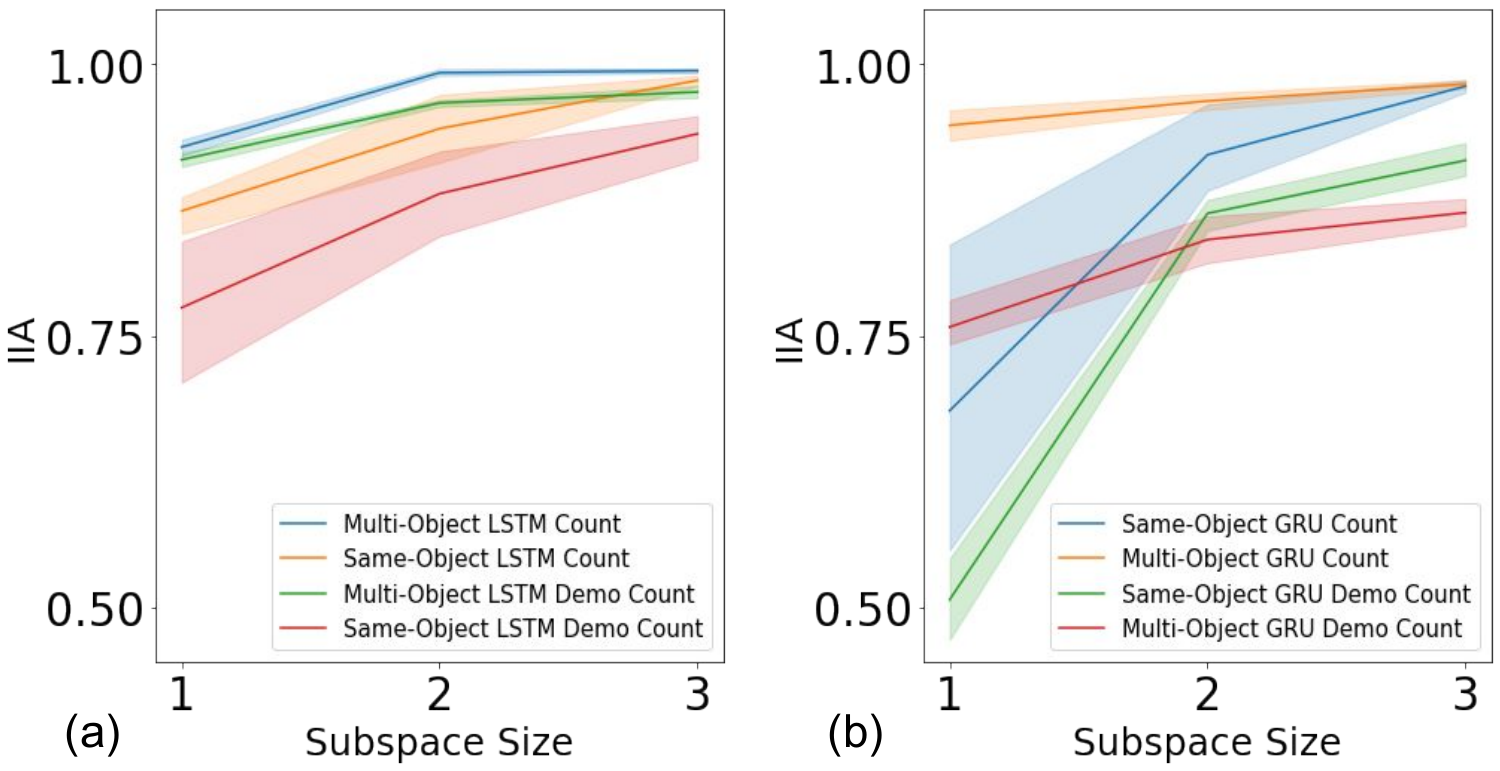}
    \caption{
        (a) An exploration of the DAS IIA on the y-axis using the Linear
        Alignment function with varying sizes of $d_{\text{var}}$ (the
        size of the intervention subspace) on the x-axis for the LSTM models.
        (b) An exploration of the DAS IIA on the y-axis using the Linear
        Alignment function with varying sizes of $d_{\text{var}}$ (the
        size of the intervention subspace) on the x-axis for the GRU models.
    }
    \label{fig:nneurons123}
\end{figure}
\begin{figure}[H]
    \centering
    \includegraphics[width=0.98\textwidth]{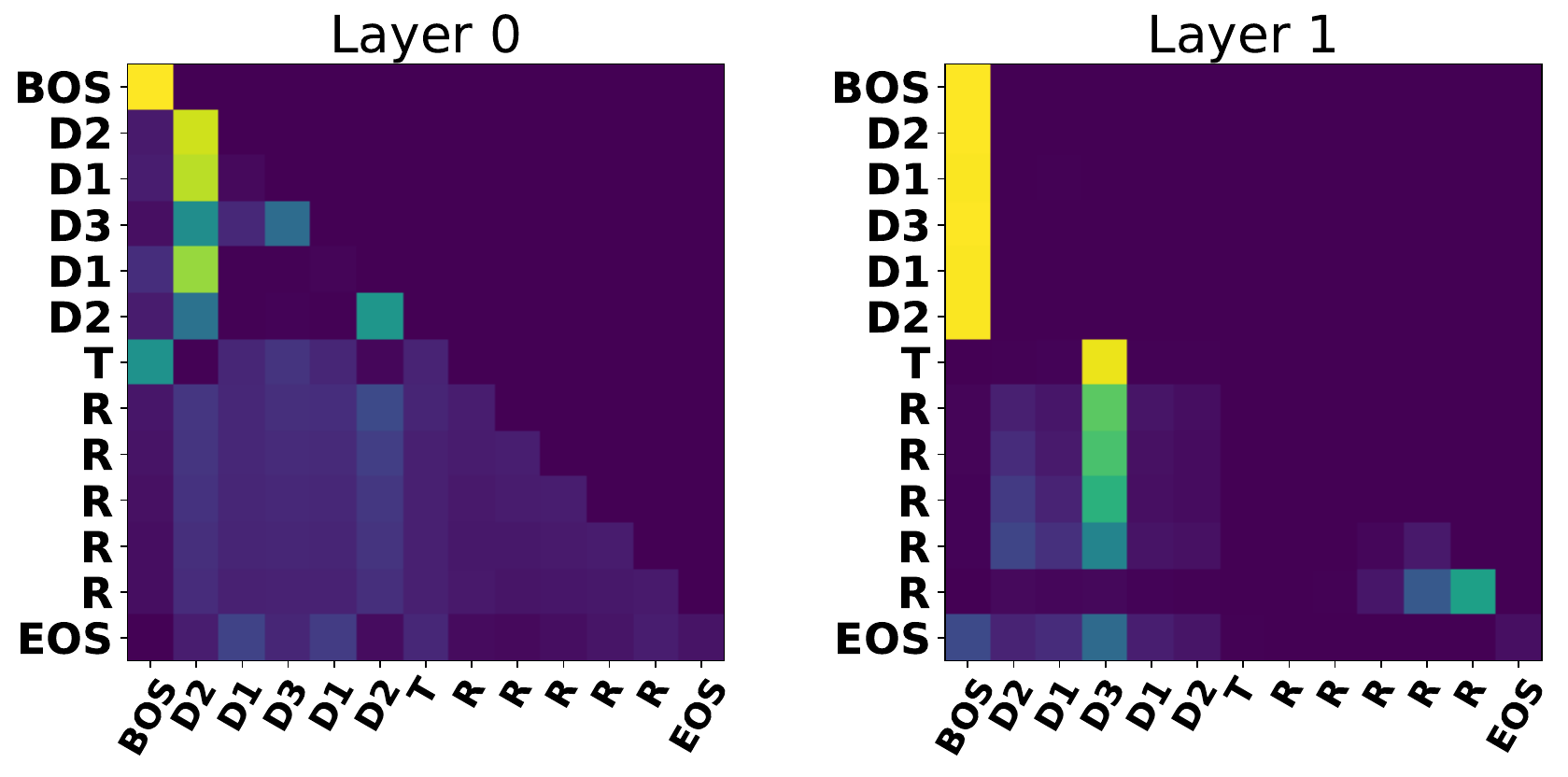}
    \caption{
    Attention weights for a single transformer with two layers
    using rotary positional encodings trained on the \multiobject\space Task.
    Queries are displayed on the vertical
    axis in order of their appearance starting at the top. Keys are displayed
    on the horizontal axis starting from the left. Queries are only able
    to attend to themselves and preceding keys.}
    \label{fig:attnrope}
\end{figure}
\begin{figure}[H]
    \centering
    \includegraphics[width=0.98\textwidth]{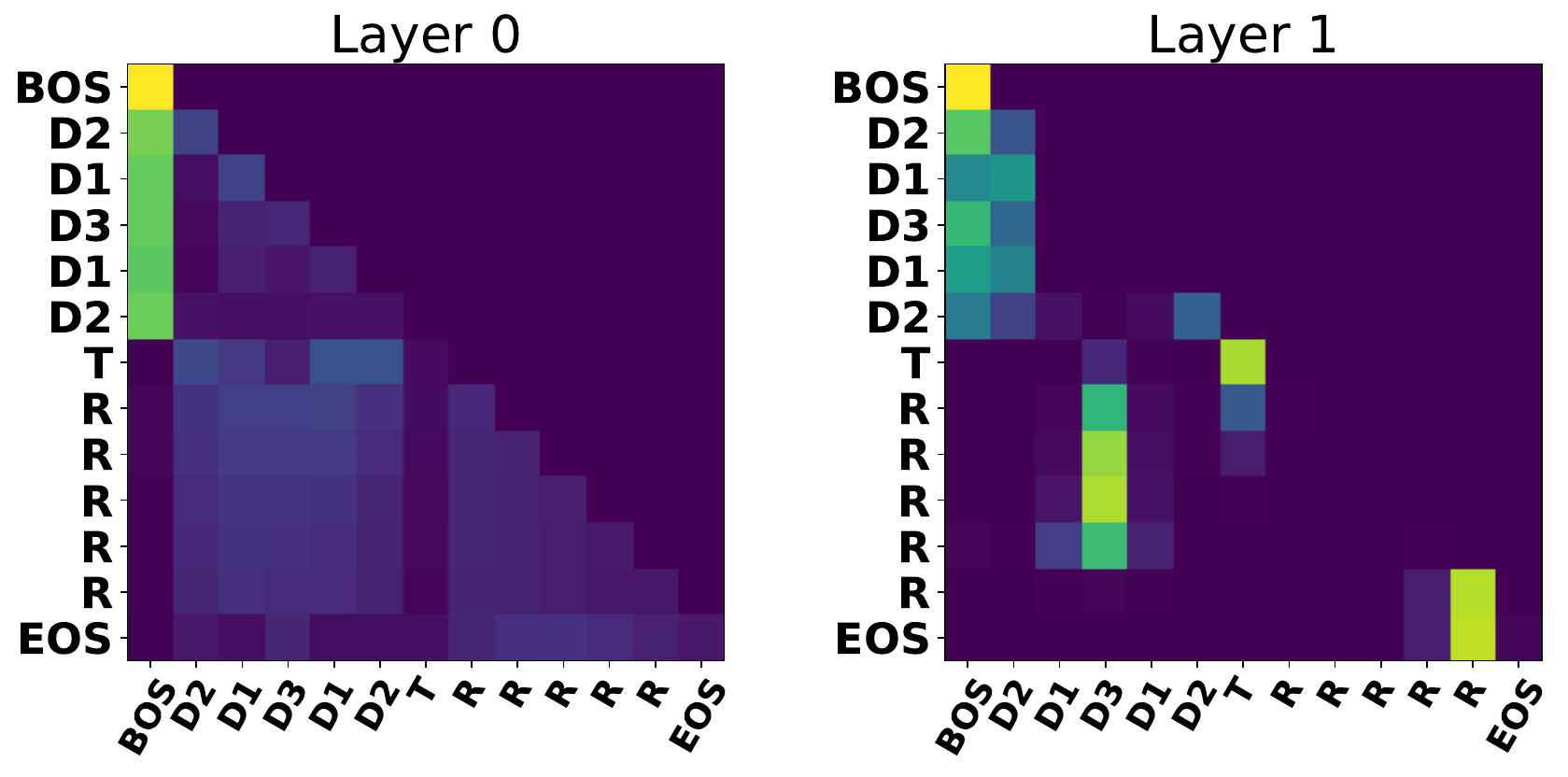}
    \caption{
    Attention weights for a single transformer with two layers
    using rotary positional encodings trained on the \varylen\space
    variant of the \multiobject\space Task. Queries are displayed on the vertical
    axis in order of their appearance starting at the top. Keys are displayed
    on the horizontal axis starting from the left. Queries are only able
    to attend to themselves and preceding keys.}
    \label{fig:attnvarylen}
\end{figure}
\begin{figure}[H]
    \centering
    \includegraphics[width=0.95\textwidth]{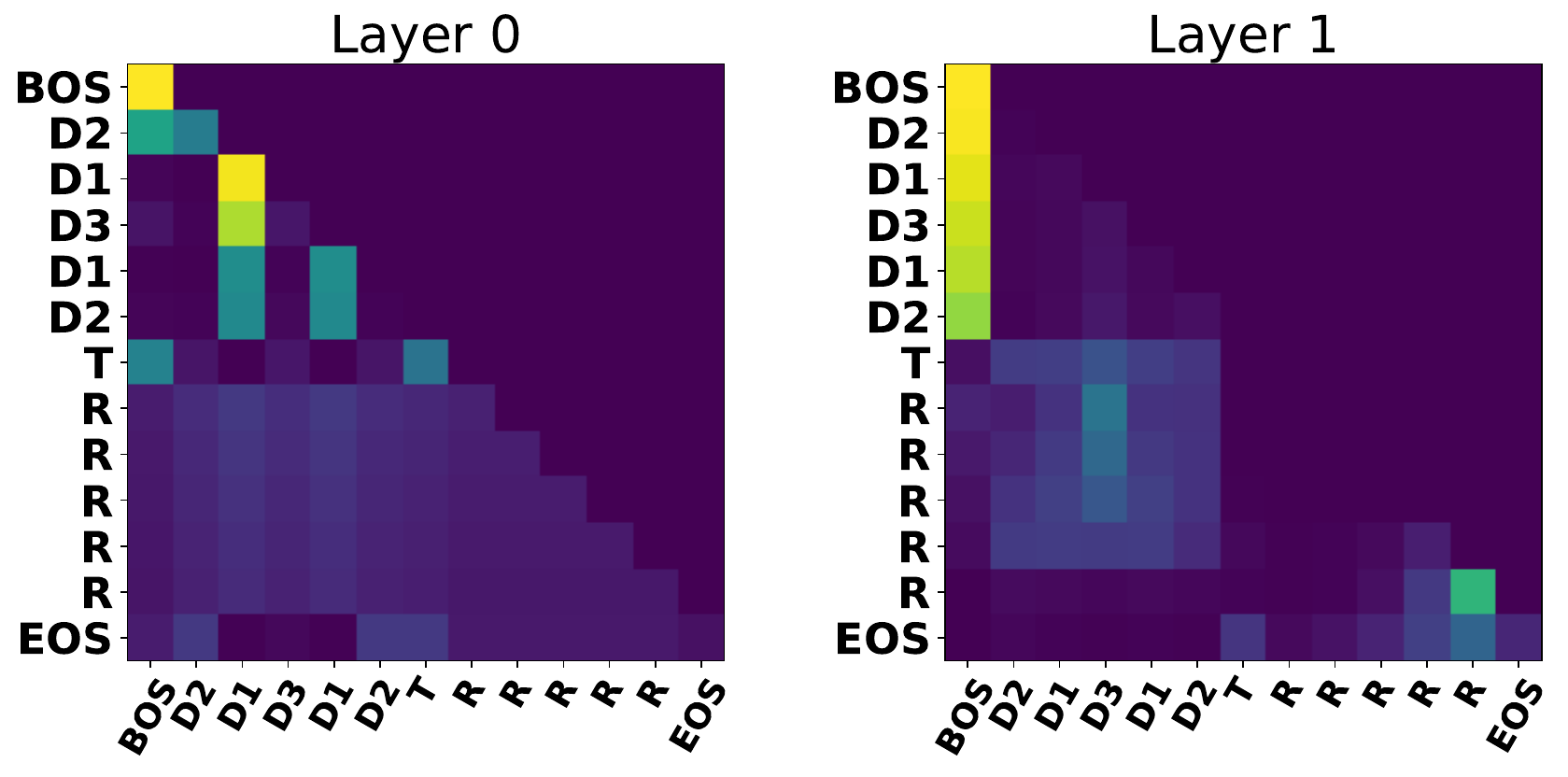}
    \caption{
    Attention weights for a single transformer model seed with two layers and
    no positional encodings (NoPE) trained on the \multiobject\space Task.
    Queries are displayed on the vertical
    axis in order of their appearance starting at the top. Keys are displayed
    on the horizontal axis starting from the left. Queries are only able
    to attend to themselves and preceding keys.}
    \label{fig:attnnpe}
\end{figure}
\begin{figure}[H]
    \centering
    \includegraphics[width=0.98\textwidth]{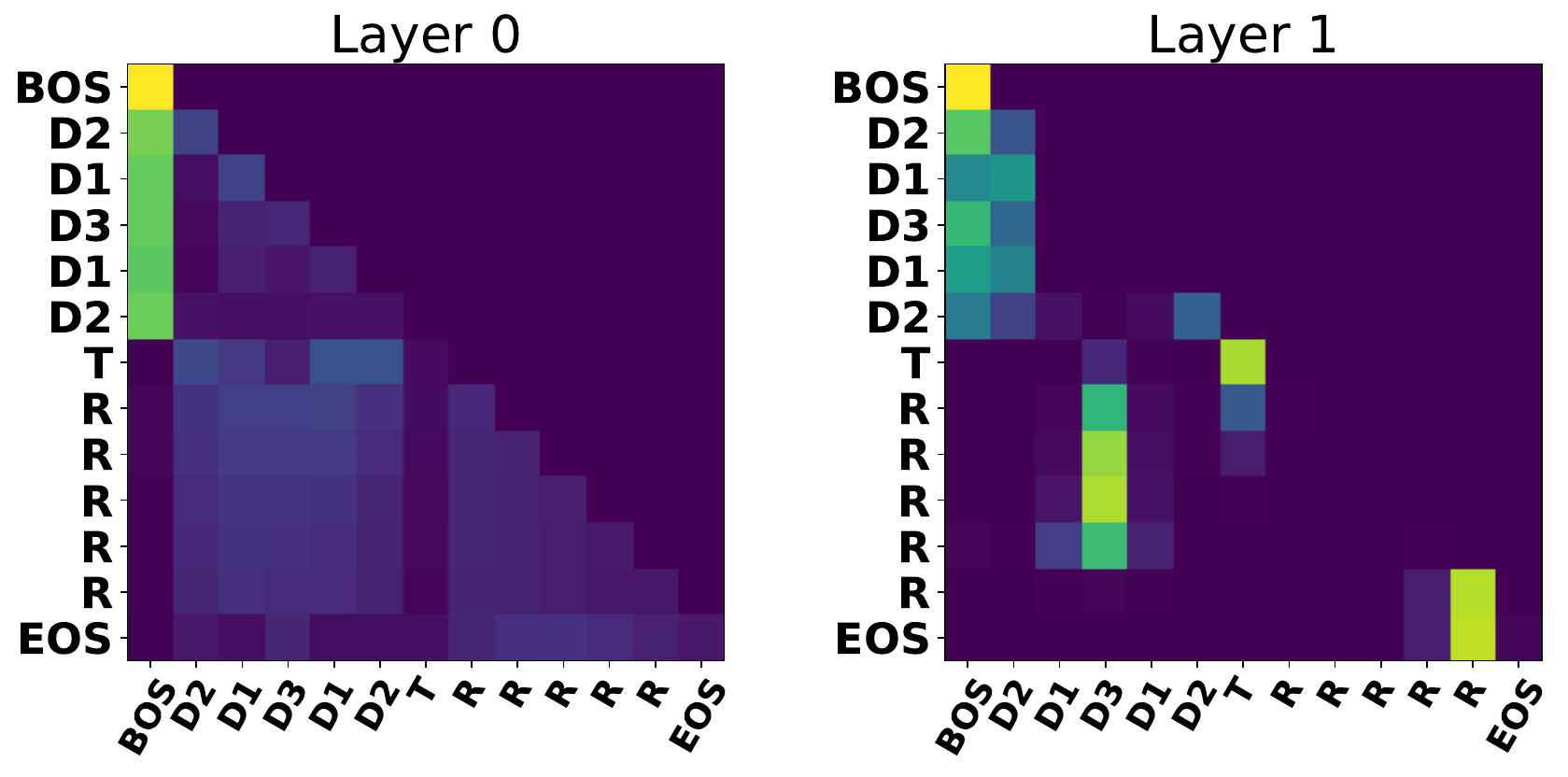}
    \caption{
    Attention weights for a single transformer with two layers
    using no positional encodings (NoPE) trained on the \varylen\space
    variant of the \multiobject\space Task. Queries are displayed on the vertical
    axis in order of their appearance starting at the top. Keys are displayed
    on the horizontal axis starting from the left. Queries are only able
    to attend to themselves and preceding keys.}
    \label{fig:attnnopevarylen}
\end{figure}

\pagebreak

\subsection{Model Details}\label{sup:modeltrain}
All artificial neural network models were implemented and trained using
PyTorch \citep{pytorch2019} on Nvidia Titan X GPUs. Unless otherwise
stated, all models used an embedding and hidden state size of
\dmodel\space dimensions. To make the token predictions, each model
used a two layer multi-layer perceptron (MLP) with GELU nonlinearities,
with a hidden layer size of 4 times the hidden state dimensionality with
50\% dropout on the hidden layer. The GRU and LSTM model variants each
consisted of a single recurrent cell followed by the output MLP. Unless
otherwise stated, the transformer architecture consisted of two layers
using Rotary positional encodings  \citep{su2023roformer}. Each model
variant used the same learning rate scheduler, which consisted of the
original transformer \citep{vaswani2017} scheduling of warmup followed
by decay. We used 100 warmup steps, a maximum learning rate of
\learnrate\space, a minimum of 1e-7, and a decay rate of 0.5. We used a
batch size of \batchsize, which caused each epoch to consist of 8
gradient update steps.

\subsection{Increment-Up Symbolic Algorithm}\label{sup:incrupsa}
\begin{figure}[h!]
    \centering
    \includegraphics[width=\textwidth]{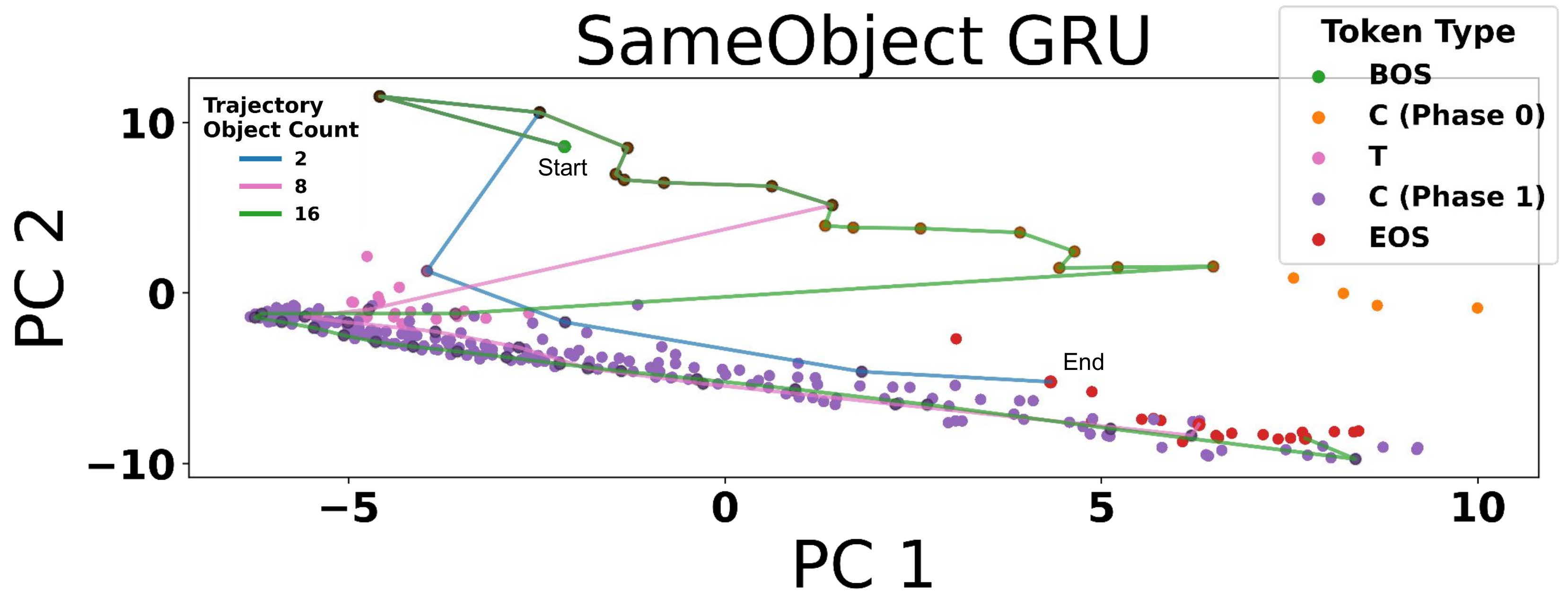}
    \caption{
    Top 2 Principal Components (PCs) of the hidden states of a single
    \sameobject\space GRU model seed where representations are taken
    from 15 sampled
    trials for each object quantity from 1 to 20 in the \sameobject\space
    task variant. Each dot represents a hidden state representation
    projected into the first 2 PCs. We plot select trial trajectories
    to exemplify how the states change for a given trial. Green points
    indicate the start of a plotted trajectory, black points indicate an
    intermediate step, and red points indicate the end of a plotted
    trajectory. The blue line plots a single trajectory from start
    to finish with a object quantity of 2. Similarly, the pink and
    green lines follow single trajectories of 8 and 16 respectively.
    }
    \label{fig:sameobjpca}
\end{figure}

In this section we explore an additional SA that we name the
\textbf{\incrupup\space Program}. This SA was in part inspired by
the PCA shown in Figure~\ref{fig:sameobjpca}. This SA has 3 variables:
the \varphase, \varprogress, and the \varincrement.
The algorithm uses the \varprogress\space variable
to track progress along a fixed interval from 0 to the maximum count
which is a preset constant. We refer to the maximum count as the
\varinterval. To solve the numeric equivalence tasks, the algorithm
increments the \varprogress\space according to an increment size which
is defined by the \varincrement\space variable.
During the demonstration phase, the algorithm initializes the
\varincrement\space to $\frac{1}{\varinterval}$. In our case the
\varinterval\space constant is 20, but we refer to it as the
\varinterval\space variable for generality. The algorithm
progressively adds this increment multiplied by the \varinterval\space
to the \varprogress\space with each successive demonstration token:
$\varprogress = \varprogress + \varincrement * \varinterval$.
Upon reaching the trigger token at the end of the
demonstration phase, the \varincrement\space is set to
$\frac{1}{\varprogress}$ and \varprogress\space is subsequently
reset to 0. The algorithm then increments the \varprogress\space by the
new value of \varincrement\space (still multiplied by the \varinterval)
until the \varprogress\space reaches a value greater than or equal to
the \varinterval. At this point, the SA returns the EOS token. See
Algorithm~\ref{alg:incrupup} for more details.

\begin{figure}[h!]
    \centering
        \centering
        \includegraphics[width=0.45\textwidth]{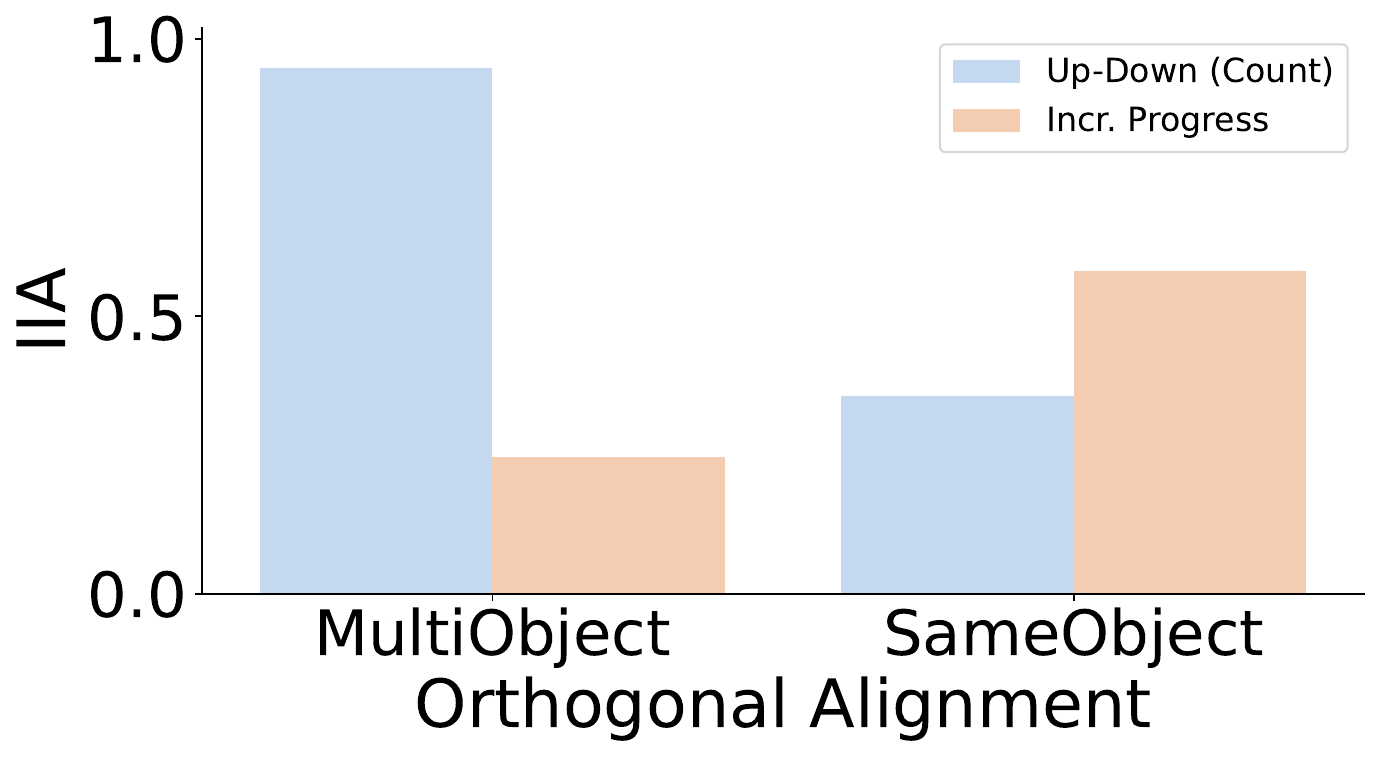}
    ~ 
        \centering
        \includegraphics[width=0.45\textwidth]{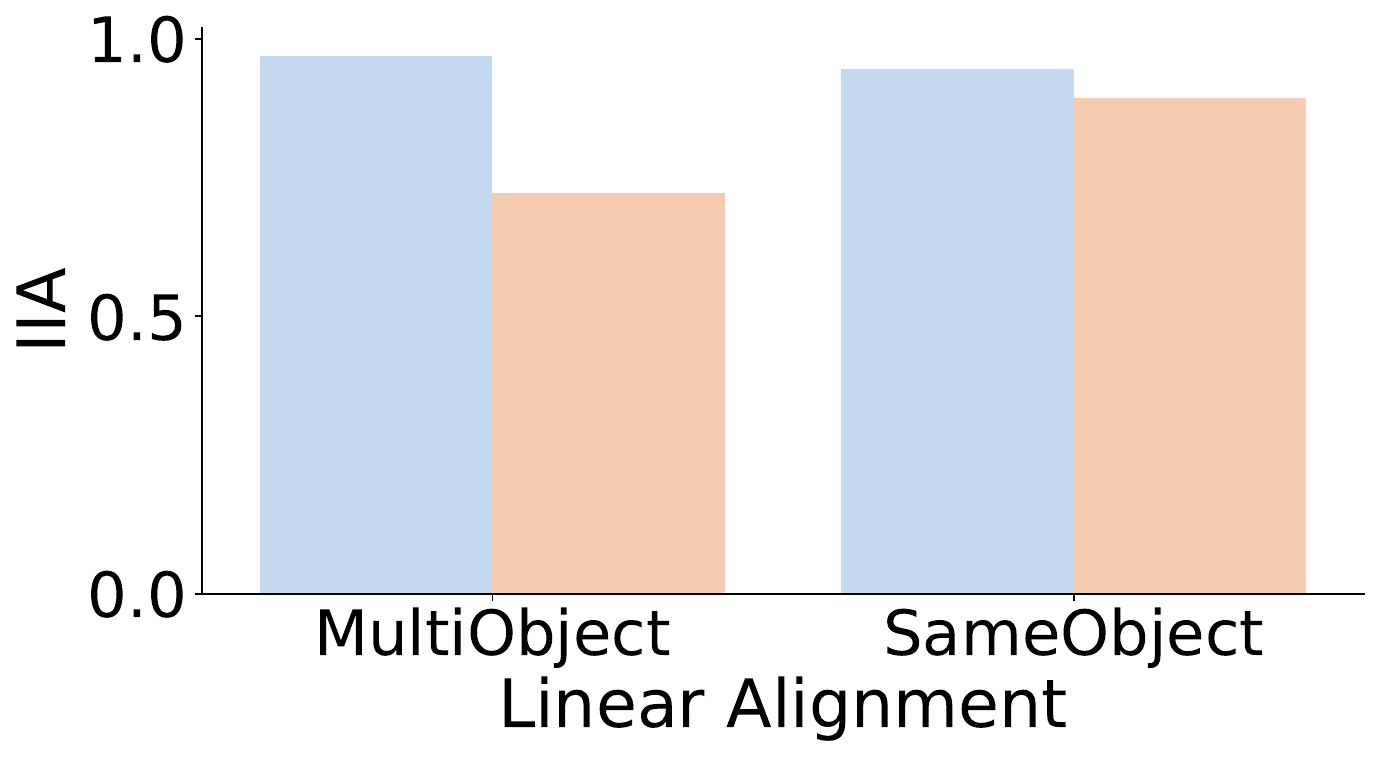}
    \caption{The Interchange intervention accuracy (IIA) for variables
    from the \countupdown\space and \incrupup\space SAs using different
    alignment functions for a single model seed for each of the
    \multiobject\space and \sameobject\space GRUs.
    The model seed is the same as that used for Figure~\ref{fig:sameobjpca}.
    The displayed IIA is for the \varcount\space variable from the
    \countupdown\space SA (Up-Down Count) and the \varprogress\space
    variable from the \incrupup\space SA (Incr. Progress). We see that
    even though the LAF raises IIA for both SAs, it does not achieve
    comparable accuracies. We use the \incrupup\space SA as a negative
    baseline experiment to demonstrate the non-triviality of high LAF
    IIA.} \label{fig:dasincrup}
\end{figure}

We compare the \incrupup\space and \countupdown\space alignments to
\multiobject\space and \sameobject\space GRUs in Figure~\ref{fig:dasincrup}.
This figure was made from the same model seed
used to produce Figure~\ref{fig:sameobjpca}. We see that IIA for the
\incrupup\space SA is relatively low for all considered models in the OAF
case. IIA is better in the LAF case, but still does not achieve the
strong accuracy of the \countupdown\space SA. We use this result to
demonstrate that LAFs cannot find good alignments with all SAs.

\subsection{DAS Training Details}\label{sup:das}
\subsubsection{Rotation Matrix Training}
To train the DAS rotation matrices, we applied PyTorch's default orthogonal
parametrization to a square matrix of the same size as the model's state
dimensionality. PyTorch creates the orthogonal matrix as the exponential of
a skew symmetric matrix. In all experiments, we selected the number of
dimensions to intervene upon as half of the dimensionality of the state.
We chose this value after an initial hyperparameter search that showed
the number of dimensions had little impact on performance (see Figure
~\ref{fig:neuronsweep}).
We sample \dasntrain\space sequence pairs for the intervention training
dataset. See Supplement~\ref{sup:intrvdata} for more details on intervention
data construction and examples. We use a learning rate of \daslr\space and a
batch size of \dasbatchsize. We removed models with performance below 99\%
to limit our DAS results to perfectly performing models thus simplifying our
interpretations of the results. We chose 99\% accuracy instead of 100\% due
to slight numerical underflow in accuracy calculations and due the fact that
half of the \varylen\space \sameobject\space models would have been dropped
due to low performance.

\subsubsection{Symbolic Program Algorithms}
\begin{algorithm}[H]
\caption{One sequence step of the \countupdown\space Program}\label{alg:updown}
\begin{algorithmic}
\State $q\gets\text{\varcount}$
\State $p\gets\text{\varphase}$
\State $y \gets \text{input\;token}$
\If{$y==\text{BOS}$} \Comment{BOS is beginning of sequence token}
    \State $q \gets 0$, $p \gets 0$
    \State $\text{return}\;\text{sample(D)}$  \Comment{sample a demo token}
\ElsIf{$y\in\text{D}$} \Comment{D is set of demo tokens}
    \State $q \gets q + 1$
    \State $\text{return}\;\text{sample(D)}$
\ElsIf{$y==\text{T}$} \Comment{T is trigger token}
    \State $p \gets 1$
\ElsIf{$y==\text{R}$} \Comment{R is response token}
    \State $q \gets q - 1$
\EndIf
\If{$(q==0)\; \& \; (p==1)$}
    \State $\text{return}\;\text{EOS}$ \Comment{EOS is end of sequence token}
\EndIf
\State $\text{return}\;\text{R}$
\end{algorithmic}
\end{algorithm}
\begin{algorithm}[H]
\caption{One sequence step of the \countupup\space Program}\label{alg:upup}
\begin{algorithmic}
\State $d\gets\text{\vardemocount}$
\State $r\gets\text{\varrespcount}$
\State $p\gets\text{\varphase}$
\State $y \gets \text{input\;token}$
\If{$y==\text{BOS}$} \Comment{BOS is beginning of sequence token}
    \State $d \gets 0$, $r \gets 0$, $p \gets 0$
    \State $\text{return}\;\text{sample(D)}$  \Comment{sample a demo token}
\ElsIf{$y\in\text{D}$} \Comment{D is set of demo tokens}
    \State $d \gets d + 1$
    \State $\text{return}\;\text{sample(D)}$
\ElsIf{$y==\text{T}$} \Comment{T is trigger token}
    \State $p \gets 1$
\ElsIf{$y==\text{R}$} \Comment{R is response token}
    \State $r \gets r + 1$
\EndIf
\If{$(d==r)\; \& \; (p==1)$}
    \State $\text{return}\;\text{EOS}$ \Comment{EOS is end of sequence token}
\EndIf
\State $\text{return}\;\text{R}$
\end{algorithmic}
\end{algorithm}
\begin{algorithm}[H]
\caption{One sequence step of the specific \distrsoln\space Program}\label{alg:distr}
\begin{algorithmic}
\State $v\gets\text{list of previous values excluding the most recent step}$
\State $\ell\gets\text{\varlastval}$ \Comment{The value of the most recent token}
\State $p\gets\text{\varphase}$ \Comment{0 indicates the demo phase, 1 is the response phase}
\State $y\gets\text{input token}$
\State
\State $v\text{.append(}\ell\text{)}$
\State $s \gets \text{SUM(}v\text{)}$
\If{$y==\text{BOS}$} \Comment{BOS is beginning of sequence token}
    \State $\ell \gets 0$, $p \gets 0$
    \State $\text{return}\;\text{sample(D)}$  \Comment{sample a demo token}
\ElsIf{$s \le 0 \text{ and } p==1$} \Comment{Sum is 0 or less in the response phase}
    \State $\text{return EOS}$ \Comment{EOS is end of sequence token}
\ElsIf{$y==\text{T or } y==\text{R}$} \Comment{T is trigger token, R is response token}
    \State $p \gets 1$
    \State $\ell \gets -1$
    \State $\text{return R}$
\ElsIf{$y\in\text{D}$}  \Comment{D is set of demo tokens}
    \State $\ell \gets 1$
\EndIf
\State
\If{$p==1$}
    \State $\text{return}\;\text{R}$
\Else
    \State $\text{return}\;\text{sample(D)}$
\EndIf
\end{algorithmic}
\end{algorithm}
\begin{algorithm}[H]
\caption{One sequence step of the \incrupup\space Program}\label{alg:incrupup}
\begin{algorithmic}
\State $m\gets\text{\varinterval}$
\State $q\gets\text{\varprogress}$
\State $p\gets\text{\varphase}$
\State $i\gets\text{\varincrement}$
\State $y \gets \text{input\;token}$
\If{$y==\text{BOS}$} \Comment{BOS is beginning of sequence token}
    \State $q \gets 0$, $p \gets 0$, $i \gets \frac{1}{m}$
    \State $\text{return}\;\text{sample(D)}$  \Comment{sample a demo token}
\ElsIf{($y\in\text{D}$ or $y==\text{R}$) and $q<m$}
    \State $q \gets q + i * m$
\ElsIf{$y==\text{T}$} \Comment{T is trigger token}
    \State $p \gets 1$
    \State $i \gets \frac{1}{q}$
    \State $q \gets 0$
\EndIf
\If{$(q\geq m)$ and $(p==1)$}
    \State $\text{return}\;\text{EOS}$ \Comment{EOS is end of sequence token}
\ElsIf{$q\geq m$ and $p==0$}
    \State $\text{return}\;\text{T}$
\ElsIf{$p==0$}
    \State $\text{return}\;\text{sample(D)}$
\Else
    \State $\text{return}\;\text{R}$
\EndIf
\end{algorithmic}
\end{algorithm}

\subsubsection{DAS Intervention Data}\label{sup:intrvdata}
Here we expand upon the intervention data used to train and test the
DAS rotation matrices. We organize this section into programs, variables,
and tasks. For each DAS training, we train a single orthonormal matrix
and only create interventions that depend on a single variable from the
corresponding program. To construct an intervention sample, we first
sample a target sequence and a source squence and a positional index
from each sequence. We limit positional indices to the demo and resp tokens.
We then compute the values of each of the variables using the symbolic
algorithm up to the positional index for both the target and source.
The value of the variable of focus is then transferred from the source
into the the target variable. We then continue the target sequence
based on the new value. When the target sequence's counterfactual
sequence begins in the demo phase, we uniformly sample the number
of demo sequence steps before placing the trigger token such that
the \varcount\space (or \vardemocount\space) does not exceed the maximum
count used in the task. We note that this makes the samples not
strictly counterfactual in the definition used in the causal
inference literature, but the desired effect is the same as the
true counterfactual comes from the same distribution.

\subsubsection{\textbf{\countupdown\space Program Examples}}

\textbf{\varcount\space Variable:}
Interventions attempt to transfer the representation corresponding
to the difference between the number of resp tokens and demo tokens.
Interventions are only performed at positional indices corresponding
to demo or resp tokens.

\begin{center}
\begin{tabular}{ c || l | l | l | l  }
\emph{\multiobject\space Examples} & \emph{1} & \emph{2}  & \emph{3}& \emph{4}              \\ 
\hline
Source Sequence & BOS $\text{D}_1$                        & BOS $\text{D}_2$ $\text{D}_1$ $\text{D}_1$ & BOS $\text{D}_2$ $\text{D}_1$ T R              & BOS $\text{D}_1$ $\text{D}_3$ T R R \\  
Target Sequence & BOS $\text{D}_3$ $\text{D}_2$           & BOS $\text{D}_2$ T R                       & BOS $\text{D}_1$ $\text{D}_2$ $\text{D}_1$ T R & BOS $\text{D}_2$         \\ 
Original Labels & $\text{D}_2$ $\text{D}_3$ T R R R R EOS & EOS                                        & R R EOS                                        & $\text{D}_2$ T R R EOS   \\
Counterfactual  & $\text{D}_2$ $\text{D}_3$ T R R R EOS   & R R R EOS                                  & R EOS                                          & $\text{D}_2$ T R EOS     \\
\hline\\
\emph{\singleobject\space Examples} & \emph{1} & \emph{2}  & \emph{3}& \emph{4}        \\ 
\hline
Source Sequence & BOS D             & BOS D D D      & BOS D D T R     & BOS D D T R R \\  
Target Sequence & BOS D D           & BOS D T R      & BOS D D D T R   & BOS D         \\ 
Original Labels & D D T R R R R EOS & EOS            & R R EOS         & D T R R EOS   \\
Counterfactual  & D D T R R R EOS   & R R R EOS      & R EOS           & D T R EOS     \\
\hline\\
\emph{\sameobject\space Examples} & \emph{1} & \emph{2}  & \emph{3}& \emph{4}        \\ 
\hline
Source Sequence & BOS C             & BOS C C C      & BOS C C T C     & BOS C C T C C \\  
Target Sequence & BOS C C           & BOS C T C      & BOS C C C T C   & BOS C         \\ 
Original Labels & C C T C C C C EOS & EOS            & C C EOS         & C T C C EOS   \\
Counterfactual  & C C T C C C EOS   & C C C EOS      & C EOS           & C T C EOS      
\end{tabular}
\end{center}

\vspace{0.75cm}
\textbf{\varphase\space Variable:}
Interventions transfer the representation corresponding
to the \varphase\space of the sequence (whether it is counting up or counting down).
Interventions are only performed at positional indices corresponding
to demo or resp tokens.

\begin{center}
\begin{tabular}{ c || l | l | l | l  }
\emph{\multiobject\space Examples} & \emph{1} & \emph{2}  & \emph{3}& \emph{4}         \\ 
\hline
Source Sequence & BOS $\text{D}_1$             & BOS $\text{D}_3$ $\text{D}_1$ $\text{D}_2$      & BOS $\text{D}_2$ $\text{D}_1$ T R     & BOS $\text{D}_2$ $\text{D}_3$ T R R \\  
Target Sequence & BOS $\text{D}_2$ $\text{D}_1$           & BOS $\text{D}_3$ T R      & BOS $\text{D}_1$ $\text{D}_3$ $\text{D}_1$ T R   & BOS $\text{D}_2$         \\ 
Original Labels & $\text{D}_3$ $\text{D}_1$ T R R R R EOS & EOS                       & R R EOS                                          & $\text{D}_1$ T R R EOS   \\
Counterfactual  & $\text{D}_3$ $\text{D}_1$ T R R R R EOS & $\text{D}_2$ T R EOS      & R R EOS         & R EOS         \\
\hline \\
\emph{\singleobject\space Examples} & \emph{1} & \emph{2}  & \emph{3}& \emph{4}        \\ 
\hline
Source Sequence & BOS D             & BOS D D D      & BOS D D T R     & BOS D D T R R \\  
Target Sequence & BOS D D           & BOS D T R      & BOS D D D T R   & BOS D         \\ 
Original Labels & D D T R R R R EOS & EOS            & R R EOS         & D T R R EOS   \\
Counterfactual  & D D T R R R R EOS & D T R EOS      & R R EOS         & R EOS         \\
\hline \\
\emph{\sameobject\space Examples} & \emph{1} & \emph{2}  & \emph{3}& \emph{4}          \\ 
\hline
Source Sequence & BOS C             & BOS C C C      & BOS C C T C     & BOS C C T C C \\  
Target Sequence & BOS C C           & BOS C T C      & BOS C C C T C   & BOS C         \\ 
Original Labels & C C T C C C C EOS & EOS            & C C EOS         & C T C C EOS   \\
Counterfactual  & C C T C C C C EOS & C T C EOS      & C C EOS         & C EOS      
\end{tabular}
\end{center}

\newpage
\subsubsection{\textbf{\countupup\space Program Examples}}

\textbf{\vardemocount\space Variable:}
Interventions attempt to transfer the representation corresponding
to the number of demo tokens in the sequence.
Interventions are only performed at positional indices corresponding
to demo or resp tokens. We remove training and evaluation samples
in which the \vardemocount\space is less than the \varrespcount\space.

\begin{center}
\begin{tabular}{ c || l | l | l | l  }
\emph{\multiobject\space Examples} & \emph{1} & \emph{2}  & \emph{3}& \emph{4}              \\ 
\hline
Source Sequence & BOS $\text{D}_1$               & BOS $\text{D}_2$  $\text{D}_3$ $\text{D}_3$      & BOS $\text{D}_2$ $\text{D}_1$ T R R            & BOS $\text{D}_1$ $\text{D}_3$ T R R \\  
Target Sequence & BOS $\text{D}_3$ $\text{D}_2$  & BOS $\text{D}_2$ $\text{D}_2$ $\text{D}_3$ T R R & BOS $\text{D}_1$ $\text{D}_2$ $\text{D}_1$ T R & BOS $\text{D}_2$         \\ 
Original Labels & T R R EOS                      & R EOS                                            & R R EOS                                        & $\text{D}_2$ T R R EOS   \\
Counterfactual  & T R EOS                        & R EOS                                            & R EOS                                          & $\text{D}_2$ T R R R EOS \\
\hline\\
\emph{\singleobject\space Examples} & \emph{1} & \emph{2}  & \emph{3}& \emph{4} \\ 
\hline
Source Sequence & BOS D             & BOS D D D          & BOS D D T R R & BOS D D T R R \\  
Target Sequence & BOS D D           & BOS D D D T R R & BOS D D D T R & BOS D         \\ 
Original Labels & T R R EOS         & R EOS           & R R EOS       & D T R R EOS   \\
Counterfactual  & T R EOS           & R EOS             & R EOS         & D T R R R EOS \\
\hline\\
\emph{\sameobject\space Examples} & \emph{1} & \emph{2}  & \emph{3}& \emph{4} \\ 
\hline
Source Sequence & BOS C             & BOS C C C          & BOS C C T C C & BOS C C T C C \\  
Target Sequence & BOS C C           & BOS C C C T C C & BOS C C C T C & BOS C         \\ 
Original Labels & T C C EOS & C EOS             & C C EOS       & C T C C EOS   \\
Counterfactual  & T C EOS   & C EOS             & C EOS         & C T C C C EOS \\
\end{tabular}
\end{center}

\vspace{0.75cm}
\textbf{\varrespcount\space Variable:}
Interventions attempt to transfer the representation corresponding
to the number of response tokens in the sequence.
Interventions are only performed at positional indices corresponding
to demo or resp tokens. We remove samples from the training and evaluation
sets that transfer a \varrespcount\space greater than the \vardemocount\space
into the response phase.

\begin{center}
\begin{tabular}{ c || l | l | l | l  }
\emph{\multiobject\space Examples} & \emph{1} & \emph{2}  & \emph{3}& \emph{4}              \\ 
\hline
Source Sequence & BOS $\text{D}_1$ $\text{D}_3$ $\text{D}_3$  & BOS $\text{D}_2$                       & BOS $\text{D}_2$ $\text{D}_1$ T R R            & BOS $\text{D}_1$ $\text{D}_3$ $\text{D}_3$ T R R R \\  
Target Sequence & BOS $\text{D}_3$ $\text{D}_2$               & BOS $\text{D}_2$ $\text{D}_2$ $\text{D}_3$ T R R & BOS $\text{D}_1$ $\text{D}_2$ $\text{D}_1$ T R & BOS $\text{D}_2$         \\ 
Original Labels & T R R EOS                                   & R EOS                                            & R R EOS                                        & $\text{D}_2$ T R R EOS   \\
Counterfactual  & T R R EOS                                   & R R R EOS                                        & R EOS                                          & $\text{D}_2$ T EOS \\
\hline\\
\emph{\singleobject\space Examples} & \emph{1} & \emph{2}  & \emph{3}& \emph{4} \\ 
\hline
Source Sequence & BOS D D D  & BOS D           & BOS D D T R R & BOS D D D T R R R \\
Target Sequence & BOS D D    & BOS D D D T R R & BOS D D D T R & BOS D         \\
Original Labels & T R R EOS  & R EOS           & R R EOS       & D T R R EOS   \\
Counterfactual  & T R R EOS  & R R R EOS       & R EOS         & D T EOS       \\
\hline\\
\emph{\sameobject\space Examples} & \emph{1} & \emph{2}  & \emph{3}& \emph{4} \\ 
\hline
Source Sequence & BOS C C C      & BOS C           & BOS C C T C C     & BOS C C C T C C C \\  
Target Sequence & BOS C C        & BOS C C C T C C & BOS C C C T C     & BOS C         \\ 
Original Labels & T C C EOS  & C EOS           & C C EOS       & C T C C EOS   \\
Counterfactual  & T C C EOS  & C C C EOS       & C EOS         & C T EOS       \\
\end{tabular}
\end{center}

\newpage
\subsubsection{\textbf{\distrsoln\space Program Examples}}\label{sup:ctxintrvdata}

\textbf{Anti-Markovian States:}
We perform these interventions directly by substituting the source hidden
state into the target hidden state without using DAS. Each intervention
examines whether the state encodes sufficient information to transfer
the NN's behavior from the source sequence into the target sequence. If the NN
uses a Markovian hidden state, then transferring the hidden state from
one position to another should result in a corresponding transfer of behavior.
In the case that the NN uses anti-Markovian states, then we would expect the
model's behavior to be unchanged at token positions that did not receive
interventions. Higher accuracies correspond to no behavioral transfer.
Interventions are only performed at positional indices corresponding to
non-terminal response tokens.

\begin{center}
\begin{tabular}{ c || l | l | l | l  }
\emph{\multiobject\space Examples} & \emph{1} & \emph{2}  & \emph{3}& \emph{4}              \\ 
\hline
Source Sequence & BOS $\text{D}_1$ $\text{D}_3$ T R R & BOS $\text{D}_2$ T R                           & BOS $\text{D}_2$ $\text{D}_1$ T R   & BOS $\text{D}_1$ $\text{D}_3$ $\text{D}_3$ T R R \\  
Target Sequence & BOS $\text{D}_3$ $\text{D}_2$ T R   & BOS $\text{D}_2$ $\text{D}_2$ $\text{D}_3$ T R & BOS $\text{D}_1$ $\text{D}_2$ T R R & BOS $\text{D}_2$ $\text{D}_1$ $\text{D}_2$ T R \\ 
Original Label  & R R EOS                             & R R R EOS                                      & EOS                                 & R R EOS   \\
Counterfactual  & R R EOS                             & R R R EOS                                      & EOS                                 & R R EOS   \\
\hline\\
\emph{\singleobject\space Examples} & \emph{1} & \emph{2}  & \emph{3}& \emph{4} \\ 
\hline
Source Sequence & BOS D D T R R & BOS DT R      & BOS D D D T R & BOS D D D T R R \\  
Target Sequence & BOS D D T R   & BOS D D D T R & BOS D D T R R & BOS D D D T R   \\ 
Original Label  & R R EOS       & R R R EOS     & EOS           & R R EOS         \\
Counterfactual  & R R EOS       & R R R EOS     & EOS           & R R EOS         \\
\hline\\
\emph{\sameobject\space Examples} & \emph{1} & \emph{2}  & \emph{3}& \emph{4} \\ 
\hline
Source Sequence & BOS C C T C C & BOS CT C      & BOS C C C T C & BOS C C C T C C \\  
Target Sequence & BOS C C T C   & BOS C C C T C & BOS C C T C C & BOS C C C T C   \\ 
Original Label  & C C EOS       & C C C EOS     & EOS           & C C EOS         \\
Counterfactual  & C C EOS       & C C C EOS     & EOS           & C C EOS         \\
\end{tabular}
\end{center}

\textbf{\varlastval\space Variable:}
These interventions attempt to transfer the representation corresponding
to the value with which the tokens contribute to the cumulative difference
between the demo and resp tokens. A value of +1 is assigned to demo tokens,
a value of -1 is assigned to resp tokens, and the algorithm stops when the
sum of the values is equal to 0 in the resp phase. Interventions are only
performed at positional indices corresponding to demo or resp tokens, and
we restrict the number of demo tokens to be at least 2 when intervening on
the demo phase. This latter restriction is to avoid cases where the cumulative
value is negative.

\begin{center}
\begin{tabular}{ c || l | l | l | l  }
\emph{\multiobject\space Examples} & \emph{1} & \emph{2}  & \emph{3}& \emph{4}              \\ 
\hline
Source Sequence & BOS $\text{D}_1$                        & BOS $\text{D}_2$                                 & BOS $\text{D}_2$ $\text{D}_1$ T R R            & BOS $\text{D}_1$ $\text{D}_3$ $\text{D}_3$ T R R R \\  
Target Sequence & BOS $\text{D}_3$ $\text{D}_2$           & BOS $\text{D}_2$ $\text{D}_2$ $\text{D}_3$ T R R & BOS $\text{D}_1$ $\text{D}_2$ $\text{D}_1$ T R & BOS $\text{D}_2$  $\text{D}_1$        \\ 
Original Labels & T R R EOS & R EOS                                            & R R EOS                                        & $\text{D}_2$ T R R R EOS   \\
Counterfactual  & T R R EOS & R R R EOS                                        & R R EOS                                        & $\text{D}_2$ T R EOS \\
\hline\\
\emph{\singleobject\space Examples} & \emph{1} & \emph{2}  & \emph{3}& \emph{4} \\ 
\hline
Source Sequence & BOS D              & BOS D           & BOS D D T R R   & BOS D D D T R R R \\
Target Sequence & BOS D D            & BOS D D D T R R & BOS D D D T R   & BOS D D         \\
Original Labels & T R R EOS  & R EOS           & R R EOS         & D T R R R EOS   \\
Counterfactual  & T R R EOS  & R R R EOS       & R R EOS         & D T R EOS       \\
\hline\\
\emph{\sameobject\space Examples} & \emph{1} & \emph{2}  & \emph{3}& \emph{4} \\ 
\hline
Source Sequence & BOS C              & BOS C           & BOS C C T C C   & BOS C C C T C C C \\
Target Sequence & BOS C C            & BOS C C C T C C & BOS C C C T C   & BOS C C       \\
Original Labels & T C C EOS  & C EOS           & C C EOS         & C T C C C EOS   \\
Counterfactual  & T C C EOS  & C C C EOS       & C C EOS         & C T C EOS       \\
\end{tabular}
\end{center}

\subsubsection{DAS Gradience Evaluation Data}\label{sup:dasevaldata}
The data used for Figures~\ref{fig:sizegradience} (c) and (d) was constructed
by sampling a single target sequence for every object count ranging from
1-\maxcount, and source sequences with object counts incrementing by 4
for each target sequence. Interventions were then constructed for each
target count, source count pair within each sequence pair. This procedure
was repeated for three times, each with a different number of steps
in the demo phase before providing the trigger token. The number of
continued demo steps was 1, 4, and 12 respectively.
    
\subsection{Context Distributed Interventions}\label{sup:sufficiency}
We detail in this section why our \distrsoln\space interchange interventions
are sufficient to demonstrate that the transformers use a solution that
re-references/recomputes the relevant information to solve the tasks at
each step in the sequence. The hidden states in Layer 1 are a bottleneck
at which a cumulative counting variable must exist if it were to use
a strategy like the \countupdown\space or \countupup\space programs.
This is because the Attention Outputs of Layer 1 are the first activations
that have had an opportunity to communicate across token positions.
This means that the representations between the Residual Stream 1 of Layer 1
up to the Residual Stream 0 of Layer 2 cannot have read a cumulative
state from the previous token position other than reading off the positional
information from the previous positional encodings. The 2-layer architecture is
then limited in that it has only one more opportunity to transfer information
between positions---the attention mechanism in Layer 2. Thus, if a hidden state
at time $t$ were to have encoded a cumulative representation of the count
that will be used by the model at time $t+1$, that cumulative representation
must exist in the activation vectors between the Residual Stream 1 in Layer 1
and the Residual Stream 0 of Layer 2. If it is using such a cumulative
representation, then when we perform a full activation swap in the Layer
1 hidden states then the resulting predictions should be influenced by
the swap.

\subsection{\varylen\space Task Variants}\label{sup:varylen}
Here we include additional tasks to prevent the transformers
with positional encodings from learning a solution that relies
on reading out positional information.
We introduce \varylen\space variants of each of the \multiobject,
\singleobject, and \sameobject\space tasks.
In the \varylen\space versions, each token in the demo phase
has a \varyprob\space probability of being sampled as a unique
"void" token type, V, that should be ignored when determining
the object quantity of the sequence. The number of demo tokens
will still be equal to the object quantity when the trigger token
is presented. We include these void tokens as a way to vary the
length of the demo phase for a given object quantity, thus breaking
correlations between positional information and object quantities.
As an example, consider the possible sequence with a object quantity
of 2: "BOS V D V V D T R R EOS".

\subsection{Anti-Markovian Proof}\label{sup:antimarkovproof}
\paragraph{Notation.} We use the following symbols throughout the theorem and proof:

\begin{itemize}
  \item $x_i$:
    the input token ID at position \(i\) in the sequence.
  \item $\text{Embed}(x_i)$:
    the embedding of token \(x_i\); this is taken to be the “residual stream” output of layer \(\ell=0\).
  \item $\ell$: 
    the layer index.  We number layers so that \(\ell=0\) is the embedding layer, and \(\ell=1,2,\dots\) are the successive self‑attention layers.
  \item $t$:
    the final position in the sequence whose cumulative state \(s_t\) we wish to encode.
  \item $h_i^\ell\in\mathbb{R}^d$:
    the residual‑stream vector at layer \(\ell\) and position \(i\).  In particular,
    \[
      h_i^0 = 
      \begin{cases}
        s_0, & i=0,\\
        \mathrm{Embed}(x_i), & i\ge1,
      \end{cases}
    \]
    and for \(\ell\ge1\),
    \[
      h_i^{\ell}
      = h_i^{\ell-1}
      \;+\;
      \mathrm{attn}_{\ell}\bigl(h_0^{\ell-1},\dots,h_i^{\ell-1}\bigr).
    \]
  \item $\mathrm{attn}_{\ell}(\cdot)$:
    the causal self‑attention update at layer \(\ell\), which may attend only to token-wise linear functions of positions \(\le i\) when computing \(h_i^\ell\).
  \item $s_i$:
    the “Markovian” cumulative state after seeing tokens up to position \(i\).  By hypothesis,
    \[
      s_i \;=\; g\bigl(s_{i-1},\,x_i\bigr),
      \quad
      s_0 \;=\; g(\,\cdot\,,\,x_0)\,,
    \]
    and we require \(h_i^\ell = s_i\) exactly when \(\ell\ge i\).
  \item $g$:
    the state‑update function \(g\!: (\text{previous state},\;\text{current token})
    \;\mapsto\;\text{new state}\).
  \item $\#\text{layers}$:
    the total number of self‑attention layers in the Transformer (excluding the embedding layer).
\end{itemize}

\textbf{Theorem:}
In a Transformer with causal self‐attention, suppose that after $\ell$ layers, position $i$ in the residual stream carries the full Markovian state
\[
s_i \;=\; g(s_{i-1}, x_i)
\]
if and only if $\ell \ge i$.  Then to encode $s_t$ at position $t$ one must have
\[
\#\text{layers} \;\ge\; t.
\]

We proceed by induction on the number of layers~$\ell$.

\medskip
\noindent\textbf{Base case} ($\ell=0$).  Layer~0 is just the embedding layer:
\[
h_0^0 = s_0 = g(\,\cdot\,, x_0),
\quad
h_i^0 = \mathrm{Embed}(x_i)\quad (i\ge1).
\]
Thus only position $0$ carries the cumulative state, and for any $t\ge1$, $\ell=0<t$ is insufficient to encode $s_t$.

\medskip
\noindent\textbf{Inductive step}.  Assume that after $\ell$ layers,
\[
\begin{cases}
h_i^\ell = s_i, & i\le \ell,\\
h_i^\ell \neq s_i, & i>\ell.
\end{cases}
\]
Consider layer $\ell+1$.  For each position $i$,
\[
h_i^{\ell+1}
\;=\;
h_i^\ell
\;+\;
\mathrm{attn}_{\ell+1}\bigl(h_0^\ell,\ldots,h_i^\ell\bigr).
\]
\begin{itemize}
  \item If $i = \ell+1$, then by causality the attention may attend only to positions $0,\dots,\ell+1$.  By the inductive hypothesis, for $j\le\ell$ we have $h_j^\ell = s_j$, and $h_{\ell+1}^\ell = f(x_{\ell+1})$ where $f$ is some function that has not seen or produced $s_{\ell+1}$. Hence the attention head can compute
  \[
    s_{\ell+1}
    \;=\;
    g\bigl(s_\ell, x_{\ell+1}\bigr)
  \]
  and add it to its residual stream, yielding $h_{\ell+1}^{\ell+1} = s_{\ell+1}$.
  \item If $i > \ell+1$, then no information can traverse more than one new position per layer, so position $i$ still does not have $s_i$.
\end{itemize}
Therefore after $\ell+1$ layers exactly positions $0,\dots,\ell+1$ carry the states $s_0,\dots,s_{\ell+1}$.  This completes the induction.

\medskip
Hence to encode the state $s_t$ at position $t$, the Transformer must have at least $t$ layers.



\subsection{Principle Components Analysis}
\begin{figure}[h!]
    \centering
    \includegraphics[width=\textwidth]{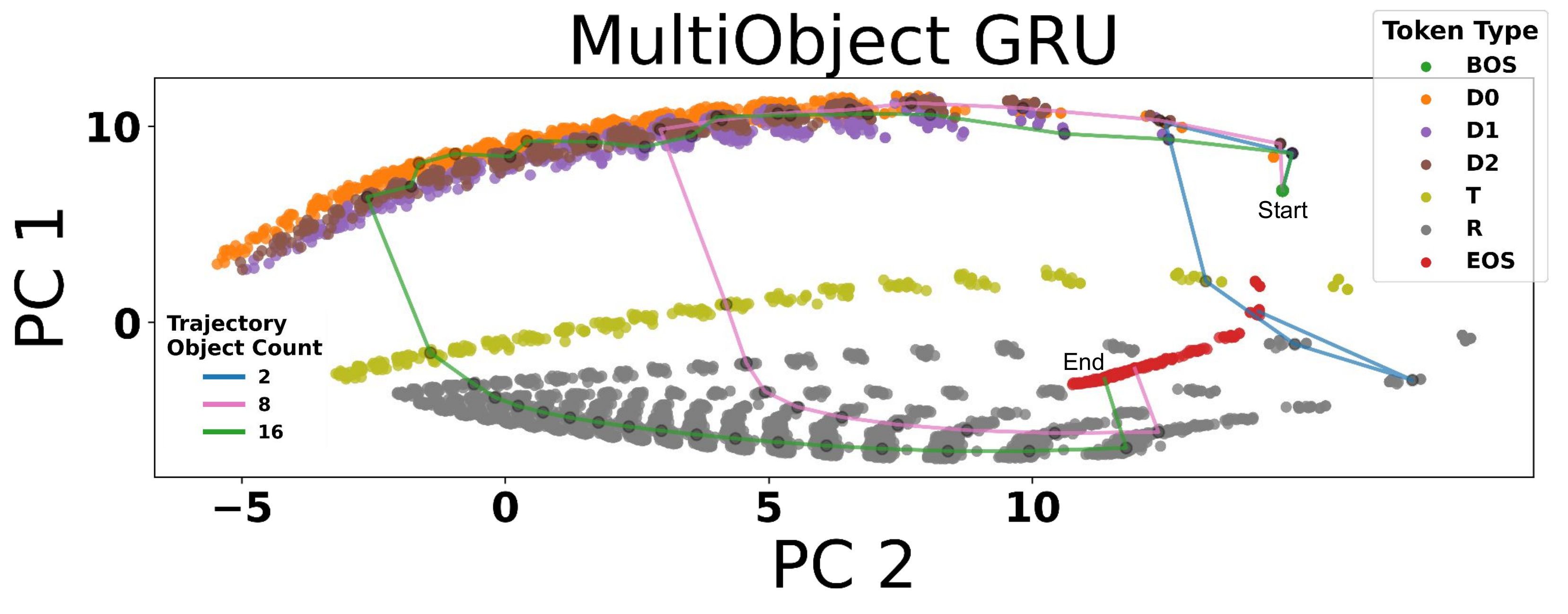}
    \caption{
    Top 2 Principal Components (PCs) of the hidden states of a single
    \multiobject\space GRU model seed where representations are taken
    from 15 sampled
    trials for each object quantity from 1 to 20 in the \multiobject\space
    task variant. Each dot represents a hidden state representation
    projected into the first 2 PCs. We plot select trial trajectories
    to exemplify how the states change for a given trial. Green points
    indicate the start of a plotted trajectory, black points indicate an
    intermediate step, and red points indicate the end of a plotted
    trajectory. The blue line plots a single trajectory from start
    to finish with a object quantity of 2. Similarly, the pink and
    green lines follow single trajectories of 8 and 16 respectively.
    }
    \label{fig:multiobjpca}
\end{figure}


\end{document}